\documentclass[11pt,letterpaper]{article}
\usepackage[margin=1in]{geometry}
\usepackage{newtxtext,newtxmath}
\usepackage{microtype}
\usepackage[table]{xcolor}
\usepackage[authoryear,round]{natbib}
\setcitestyle{authoryear,round,citesep={;},aysep={,},yysep={;}}
\usepackage{booktabs,graphicx,multirow,wrapfig,array,tabularx}
\usepackage[normalem]{ulem}
\usepackage{enumitem,float}
\usepackage[most]{tcolorbox}
\usepackage[font=small,labelfont=bf]{caption}
\usepackage{url}
\usepackage{hyperref}

\usepackage{amsmath,amsfonts,bm}

\def\eqref#1{equation~\ref{#1}}

\def\1{\bm{1}}

\DeclareMathAlphabet{\mathsfit}{\encodingdefault}{\sfdefault}{m}{sl}
\SetMathAlphabet{\mathsfit}{bold}{\encodingdefault}{\sfdefault}{bx}{n}

\newcolumntype{Y}{>{\raggedright\arraybackslash}X}

\definecolor{preprintblue}{RGB}{30,66,112}
\newcommand{\papertitle}{Look Before You Select: Rethinking Vocabulary Sparsification in On-Policy Distillation}
\hypersetup{
  colorlinks=true,
  linkcolor=preprintblue,
  citecolor=preprintblue,
  urlcolor=preprintblue,
  pdftitle={Look Before You Select: Rethinking Vocabulary Sparsification in On-Policy Distillation},
  pdfauthor={Yongliang Miao, Shuang Liu, Yanguang Liu, Yandong Bai, Mengnan Du},
}

\title{\papertitle}
\date{}

\author{
  Yongliang Miao$^{1}$ \quad
  Shuang Liu$^{2}$ \quad
  Yanguang Liu$^{3}$ \quad
  Yandong Bai$^{4}$ \quad
   Mengnan Du$^{1}$\textsuperscript{†}\\
  $^{1}$The Chinese University of Hong Kong, Shenzhen \quad
  $^{2}$Carnegie Mellon University \\
  $^{3}$New Jersey Institute of Technology \quad
  $^{4}$Kuaishou \\
  \texttt{r130026108@gmail.com} \quad
  \texttt{mengnandu@cuhk.edu.cn}\\
  \textsuperscript{†}Corresponding author
}

\begin{document}

\maketitle

\begin{abstract}
On-policy distillation (OPD) uses teacher correction on student-generated responses. Full-vocabulary correction can provide important corrections even for tokens that the student assigns low probability, but backpropagating through all token logits becomes memory-intensive for long sequences. Existing memory-saving approaches estimate corrections from sampled tokens or restrict supervision to the student's TopK tokens, introducing sampling noise or changing the full-vocabulary correction. We introduce \textbf{SparseOPD}, which uses full-vocabulary teacher correction to determine which corrections matter before selecting the token logits to differentiate. SparseOPD first constructs the full-vocabulary correction without retaining its backward graph, then selects tokens by correction magnitude rather than student probability. Signed residual compensation preserves the total promoting and suppressing correction mass, while correction-aware budget allocation distributes the sparse support across positions. Finally, the update backpropagates only through the selected token logits. Across six task--scale settings spanning mathematics, chemistry QA, and multimodal reasoning, SparseOPD outperforms Sampled Token and TopK in task-average accuracy and matches or exceeds Full Vocabulary. Gradient cosine similarity reaches 99\% on 4B mathematics, while 8K full-parameter profiling shows 70.5\% lower backward memory.

\end{abstract}

\raggedbottom
\section{Introduction}
\label{sec:introduction}

On-policy distillation (OPD) trains a student language model using teacher
correction on responses generated by the student itself~\citep{agarwal2024onpolicy,ko2024distillm,ye2025gad,song2026survey,li2026rethinkopd}.
Full-vocabulary OPD compares the complete teacher and student distributions at
each prediction position~\citep{gu2024minillm,zhong2024revisiting,zhang2024dualspace,wu2025akl,zhang2025aligndistil},
providing a token-specific correction over the entire vocabulary.
Such correction can capture important corrections beyond the student's most
likely tokens, but directly backpropagating through the full vocabulary is
memory-intensive, especially for long reasoning sequences.
This creates an apparent trade-off: preserving full-vocabulary correction seems
to require backpropagating through the full vocabulary.

\begin{figure}[!ht]
  \centering
  \graphicspath{{figures/}{iclr-2027-style-files/iclr2027/figures/}}
  \includegraphics[width=0.735\linewidth]{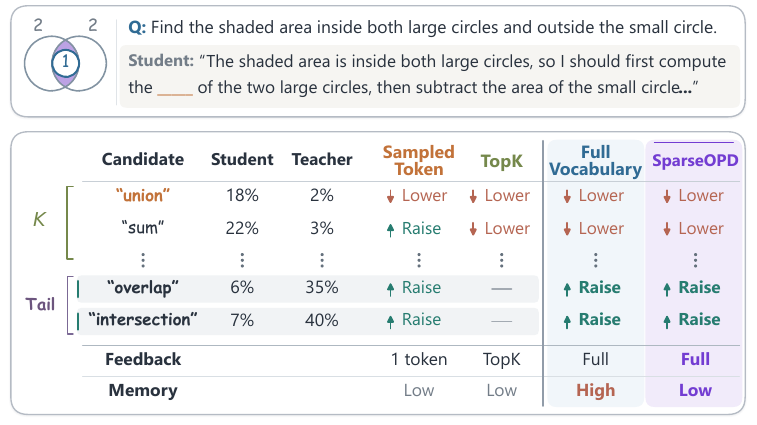}
  \setlength{\abovecaptionskip}{5pt}
  \setlength{\belowcaptionskip}{0pt}
  \caption{\textbf{Full-vocabulary correction need not require full-vocabulary differentiation.}
Arrows show logit corrections in a constructed multimodal QA example.
Full Vocabulary suppresses ``union'' and ``sum'' and promotes
``overlap'' and ``intersection,'' at high memory cost.
Sampled Token incorrectly promotes ``sum,'' while TopK provides
no direct correction to ``overlap'' or ``intersection.''
SparseOPD preserves the important full-vocabulary corrections with low memory use.}
  \label{fig:intro_correction}
\end{figure}

Existing memory-saving paradigms reduce this cost by changing the correction, typically through sampled-token estimation or
TopK support restriction~\citep{anshumann2025sparse,huang2026tailaware}. Sampled Token replaces the deterministic full-vocabulary correction with an estimate based on a single sampled token; although unbiased at a fixed prefix, the resulting update can depend strongly on which token happens to be sampled.
TopK conditions the objective on a support formed by the student's most probable tokens, leaving tail tokens that may require large corrections outside the support and changing the correction even within the retained support.
As illustrated in Figure~\ref{fig:intro_correction}, a sampled action can incorrectly promote “sum”, while a TopK support can omit the teacher-preferred tail tokens “overlap” and “intersection”.


Motivated by these limitations, we ask:
\emph{\uline{Can we construct updates from full-vocabulary teacher correction while backpropagating through the logits of only a sparse set of tokens?}} To this end, we introduce \textbf{SparseOPD}, which uses full-vocabulary correction while backpropagating only through the logits of selected tokens. SparseOPD first constructs the full-vocabulary correction
without retaining its backward graph, and then converts it into a fixed sparse
update. Because the complete correction is available before sparsification, SparseOPD selects tokens based on their corrections rather than student probability, allowing low-probability tokens to be retained when their corrections are large. The discarded positive and negative correction mass is then redistributed among retained corrections of the same sign, preserving the total positive and negative correction mass of the dense update. The resulting update therefore backpropagates only through the logits of the selected tokens.

Empirically, we validate SparseOPD on mathematical reasoning, chemistry
question answering, and multimodal mathematical reasoning.
With an average support budget of $K$ entries per position, SparseOPD outperforms
Sampled Token and TopK in all six task--scale settings and matches or
exceeds Full Vocabulary (Figure~\ref{fig:intro_results}).
Full-parameter profiling of the 4B model shows 70.5\% lower peak
backward memory than Full Vocabulary at 8K sequence length, with larger relative savings
across longer measured sequences.
Gradient analysis in the 4B mathematical reasoning setting further shows
a training-averaged cosine similarity of 0.989 to the Full Vocabulary
gradient.

Our contributions are as follows:
\begin{itemize}[leftmargin=10pt, topsep=-2pt, itemsep=1pt, partopsep=1pt, parsep=1pt]
    \item We rethink vocabulary sparsification in OPD by treating the
    full-vocabulary logit correction as the reference for sparse updates,
    showing that Sampled Token introduces sampling variance while TopK omits
    tail corrections and changes corrections through restricted-support
    conditioning.
   \item 
  We introduce SparseOPD, which constructs full-vocabulary corrections before converting them into sparse updates using signed correction selection, residual compensation, and budget allocation across the batch.

  \item   Across mathematical reasoning, chemistry QA, and multimodal mathematical
  reasoning, SparseOPD achieves higher task-average accuracy than
  Sampled Token and TopK. Complementary profiling and gradient measurements
  on the 4B mathematical reasoning setting show substantially lower backward
  memory and close agreement with the Full Vocabulary gradient.
\end{itemize}

\section{Rethinking Vocabulary Sparsification}
\label{sec:rethinking}

\subsection{Preliminaries: Full-Vocabulary OPD}
\label{sec:background}

Let $\pi_\theta$ be a student with trainable parameters $\theta$
and $\pi_T$ a frozen teacher, sharing a vocabulary $\mathcal V$.
On-policy distillation (OPD) obtains conditioning prefixes from the
student's own generations~\citep{agarwal2024onpolicy}.
For a prompt $x\sim\mathcal D$ from the training distribution, let
$y\sim\pi_\theta(\cdot\mid x)$ denote the student rollout and
$s_t=(x,y_{<t})$ its prefix at position $t$.
Both models are evaluated on the same prefix, yielding
$p_{t,v}=\pi_\theta(v\mid s_t)$ and
$q_{t,v}=\pi_T(v\mid s_t)$ for each token $v\in\mathcal V$.
Training averages per-position losses over valid completion tokens and
treats the sampled rollout as observed data, without differentiating
through discrete sampling. Within this common setting, the paired
distributions can yield training signals with different computational
costs and statistical properties.

\paragraph{Full-vocabulary OPD.}
The full-vocabulary formulation compares the complete distributions at
each prefix using reverse Kullback--Leibler (KL) divergence. Let $z_{t,v}$
denote the student logit and $r_{t,v}=\log(p_{t,v}/q_{t,v})$ the student-to-teacher log-ratio. The per-position loss and its logit-gradient
coefficients are
\begin{equation}
D_t=\mathrm{KL}(p_t\Vert q_t)
   =\sum_{v\in\mathcal V}p_{t,v}r_{t,v},
\qquad
C_{t,v}:=\frac{\partial D_t}{\partial z_{t,v}}
   =p_{t,v}(r_{t,v}-D_t).
\label{eq:opd_coefficient}
\end{equation}

The vector $C_t=(C_{t,v})_{v\in\mathcal V}$ specifies a deterministic,
token-specific correction over the entire vocabulary.
Positive coefficients suppress the corresponding logits under gradient
descent, while negative coefficients promote them. The coefficients are
zero-sum, $\sum_{v\in\mathcal V} C_{t,v}=0$, so the total suppressive and
promotive correction masses are equal. The chain rule maps these
coefficients to the model's parameter gradient.
Directly differentiating this loss retains vocabulary-sized intermediates
for backward, with memory cost growing with both vocabulary size and the
number of valid positions.

\subsection{What Existing Sparsification Changes}
\label{sec:existing_sparsification}

\paragraph{Sampled-token OPD.}
One way to avoid evaluating every token-specific discrepancy is to reuse
the generated action $a_t=y_t\sim p_t$. Its single-action coefficient
estimator satisfies
\begin{equation}
\widehat C^{\mathrm{sample}}_{t,v}
=r_{t,a_t}\bigl(\mathbf 1\{v=a_t\}-p_{t,v}\bigr),
\qquad
\mathbb E_{a_t\sim p_t}\!\left[\widehat C^{\mathrm{sample}}_{t,v}\right]
=C_{t,v}.
\label{eq:sampled_opd}
\end{equation}
Thus the full-vocabulary correction is recovered in expectation at a fixed
prefix. In one realization, however, every coordinate depends on the
single observed ratio $r_{t,a_t}$. The $-p_{t,v}$ term couples unsampled
logits to that action; it does not evaluate their own teacher--student
discrepancies. Their token-specific corrections emerge through the sampling
expectation, introducing sampling variance.
\paragraph{TopK OPD.}
An alternative restricts the objective to the student's
highest-probability tokens. Let $S_t\subset\mathcal V$ contain the $K$
largest student probabilities, with indices fixed during
backpropagation. Define the support masses
$Z_t^p=\sum_{v\in S_t}p_{t,v}$ and
$Z_t^q=\sum_{v\in S_t}q_{t,v}$, and distributions
$\bar p_{t,v}^S=p_{t,v}/Z_t^p$ and
$\bar q_{t,v}^S=q_{t,v}/Z_t^q$ for $v\in S_t$.
TopK minimizes $D_t^S=\mathrm{KL}(\bar p_t^S\Vert\bar q_t^S)$.

Renormalization changes which discrepancies the objective measures.
Let $D_t^{S^c}$ be the analogous conditional KL on
$S_t^c=\mathcal V\setminus S_t$, and let $D_t^{\mathrm{mass}}$ be the
binary KL between the support probabilities $Z_t^p$ and $Z_t^q$.
Separating support membership from the distributions within each part gives
\begin{equation}
D_t=
\underbrace{D_t^{\mathrm{mass}}}_{\text{support--tail mass}}
+\underbrace{Z_t^pD_t^S}_{\text{support shape}}
+\underbrace{(1-Z_t^p)D_t^{S^c}}_{\text{tail shape}}.
\label{eq:topk_decomposition}
\end{equation}
TopK optimizes only the conditional shape $D_t^S$, leaving out the
support--tail mass discrepancy, the tail shape, and the support
weight. This restriction also appears directly in its logit coefficients:
\begin{equation}
C_{t,v}^{\mathrm{TopK}}
=
\bar p_{t,v}^S
\left(\log\frac{\bar p_{t,v}^S}{\bar q_{t,v}^S}-D_t^S\right)
\;(v\in S_t),\qquad
0\;(v\notin S_t).
\label{eq:topk_coefficient}
\end{equation}
Tail coordinates receive no direct correction, and the retained
coefficients are derived from a conditional objective rather than by
truncating $C_t$.

Full-vocabulary OPD provides a complete reference correction, but its
direct realization is memory-intensive. Sampling and support conditioning
limit the token-specific feedback in different ways, yielding a noisy
realization or a restricted objective.

\subsection{Full-Vocabulary Correction with Sparse Execution}
\label{sec:task_definition}

Despite these different mechanisms, both memory-saving paradigms alter the
reference correction before deciding which token logits to backpropagate
through: Sampled Token estimates it from a sampled action, while TopK defines
it only after restricting the vocabulary support. This suggests a different
order of operations: first construct the full-vocabulary correction, and only
then decide which token logits to retain for backpropagation. The task is
therefore to approximate the full-vocabulary correction as faithfully as
possible under a shared budget on token logits for backpropagation, while
retaining access to the complete teacher and student distributions. This
reframes vocabulary sparsification as correction compression rather than simple
support restriction.

\begin{center}
\setlength{\fboxsep}{7pt}
\setlength{\fboxrule}{0.55pt}
\fcolorbox{black!65}{violet!3}{%
\begin{minipage}{\dimexpr\linewidth-2\fboxsep-2\fboxrule\relax}
\textbf{Target: Full-vocabulary correction with sparse execution.}
\par\smallskip
Let $i=1,\ldots,N$ index the valid completion positions in one optimizer
update, with full-vocabulary reference coefficients $C_i$.
Given an average support budget $K$, use the complete $C_i$ to construct
supports $S_i\subseteq\mathcal V$ and coefficients $\widetilde C_i$
supported on $S_i$, subject to
\[
\sum_{i=1}^{N}|S_i|=KN.
\]
The goal is to approximate $C_i$ in coefficient space while restricting
output-head differentiation to the selected entries $S_i$.
\end{minipage}}
\end{center}

\section{SparseOPD}
\label{sec:method}
\begin{figure}[t]
  \centering
  \graphicspath{{figures/}{iclr-2027-style-files/iclr2027/figures/}}
  \includegraphics[width=\linewidth]{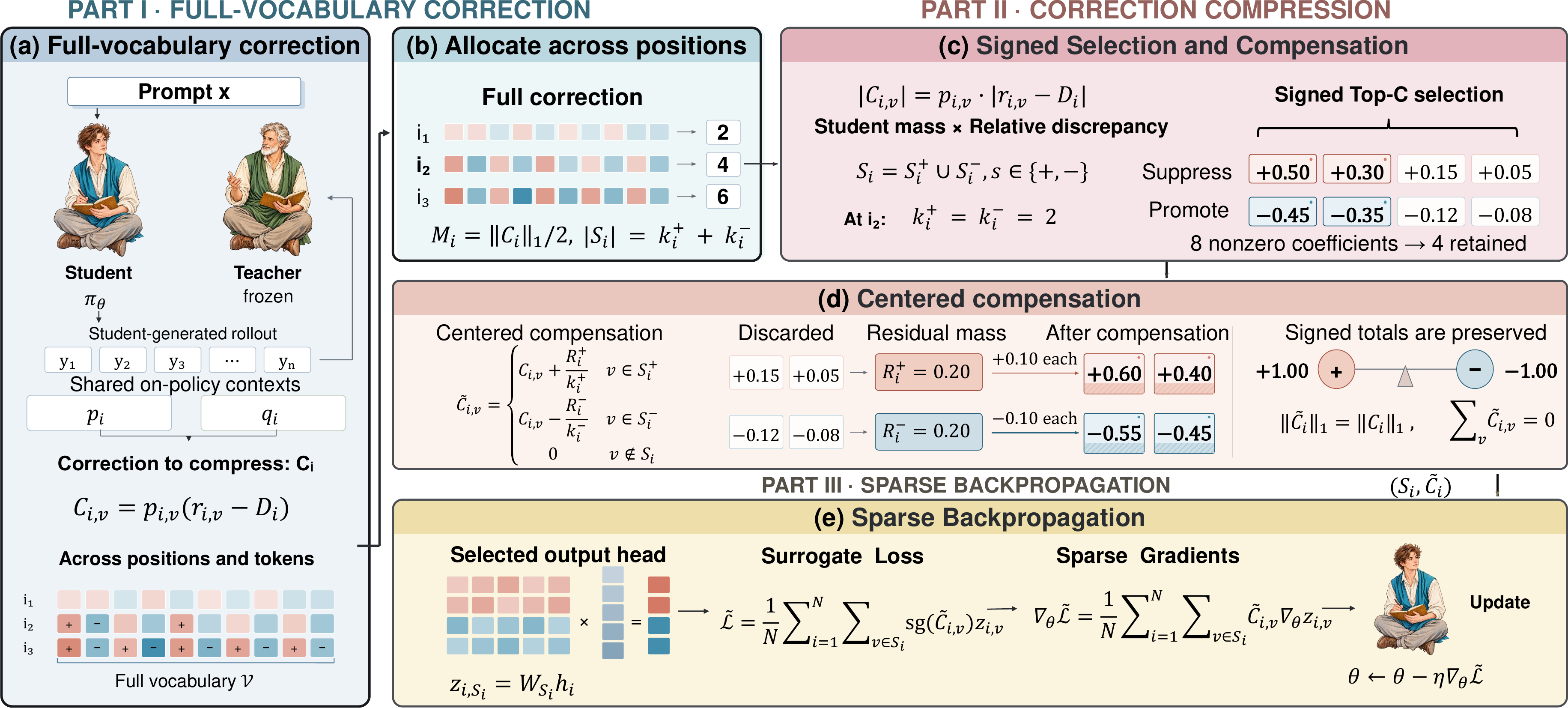}
  \setlength{\abovecaptionskip}{3pt}
  \setlength{\belowcaptionskip}{0pt}
  \vspace{-6pt}
\caption{
\textbf{Overview of SparseOPD.}
\textbf{(a)} Full-vocabulary correction
$C_{i,v}=p_{i,v}(r_{i,v}-D_i)$.
\textbf{(b)} Correction mass $M_i=\lVert C_i\rVert_1/2$ allocates support.
\textbf{(c)} Signed Top-$C$ selects $S_i=S_i^+\cup S_i^-$.
\textbf{(d)} Centered compensation yields $\widetilde C_i$ while preserving
signed totals.
\textbf{(e)} Sparse backpropagation passes only through selected logits
$z_{i,S_i}=W_{S_i}h_i$.
}
  \label{fig:pipeline}
\end{figure}

Following the reformulation of vocabulary sparsification as correction
compression in Section~\ref{sec:rethinking}, we propose \textbf{SparseOPD}, which takes the full-vocabulary correction as the
reference and backpropagates only through a sparse set of token logits. SparseOPD first constructs the full-vocabulary reference coefficients
without building an autograd graph, and then compresses this correction into
a sparse execution plan that approximates the full-vocabulary update with far
fewer differentiable entries. With the full correction available before
support selection, the compression is guided by the correction itself. SparseOPD consists of three stages:
(1) batch-global budget allocation distributes the shared support budget
according to correction mass;
(2) signed selection and compensation retains large-magnitude coefficients
within each sign and redistributes discarded mass while preserving the signed totals;
(3) Sparse Backpropagation executes the resulting compressed
correction through selected output-head rows.
Figure~\ref{fig:pipeline} provides an overview of our method.

\subsection{Batch-global Budget Allocation}
\label{sec:budget_allocation}

We first compute the full-vocabulary coefficients $C_i$ at each valid
position using Equation~(\ref{eq:opd_coefficient}), without building an
autograd graph for this computation.
Uniform support may over-allocate capacity to positions with weak
teacher--student disagreement and under-allocate it where corrections are
stronger. We
therefore share the $KN$ budget across positions according to their total
coefficient magnitude. By the zero-sum property of $C_i$, its suppressive
and promotive corrections carry equal total mass:
\begin{equation}
M_i:=\sum_{v:C_{i,v}>0}C_{i,v}
=\sum_{v:C_{i,v}<0}(-C_{i,v})
=\frac12\|C_i\|_1.
\label{eq:signed_mass}
\end{equation}
We use the common mass $M_i$ as a position-level measure of correction
strength, while the two signs distinguish suppressive and promotive
corrections.

Each active sign needs at least one retained entry for the compensation
below. For $s\in\{+,-\}$, let $\ell_i^s=\mathbf1\{M_i>0\}$ be this
minimum and $u_i^s\ge\ell_i^s$ a prescribed capacity, no larger than the
number of coefficients with that sign. Assuming the total budget is feasible,
$\sum_{i,s}\ell_i^s\le KN\le\sum_{i,s}u_i^s$, let $\alpha\ge0$ denote
a global allocation scale. We distribute the remaining budget in proportion
to $M_i$ until each group reaches its capacity:
\begin{equation}
\bar k_i^s=\ell_i^s+
\min\{u_i^s-\ell_i^s,\,\alpha M_i\},
\qquad
\sum_{i,s}\bar k_i^s=KN.
\label{eq:global_allocation}
\end{equation}
We choose $\alpha$ to meet the shared budget and round $\bar k_i^s$ to
integer cardinalities $k_i^s$ while preserving both the bounds and
$\sum_{i,s}k_i^s=KN$. Thus positions share capacity according to correction
demand rather than receiving the same support size independently.

\subsection{Signed Selection and Compensation}
\label{sec:coefficient_compression}

With the support sizes fixed, we prioritize coordinates whose
removal most distorts $C_i$. Student probability alone does not measure this
importance, because the coefficient magnitude factorizes as
\begin{equation}
|C_{i,v}|=
\underbrace{p_{i,v}}_{\text{student mass}}
\;\cdot\;
\underbrace{|r_{i,v}-D_i|}_{\text{relative discrepancy}}.
\label{eq:coefficient_magnitude}
\end{equation}
A token's update importance depends on both its student mass and its
discrepancy relative to the KL baseline. A high-probability token can
require little correction, while a tail token can matter when its
relative discrepancy is large.

For a fixed support size $k$, the direct truncation error is
$\sum_{v\notin S}C_{i,v}^2$. Selecting the $k$ largest magnitudes
$|C_{i,v}|$ minimizes this error. This coefficient ranking, which we call
Top-$C$, determines which individual entries are most costly to discard.

With the cardinalities fixed, signed Top-$C$ retains the $k_i^+$ largest
positive coefficients and the $k_i^-$ largest-magnitude negative
coefficients, forming $S_i^+$ and $S_i^-$ with union $S_i$.

\paragraph{Centered compensation.}
Simply truncating to these supports would discard correction mass and can
destroy the zero-sum structure of $C_i$. The missing masses are
$R_i^+=M_i-\sum_{v\in S_i^+}C_{i,v}$ and
$R_i^-=M_i-\sum_{v\in S_i^-}(-C_{i,v})$.
To preserve the total correction in each direction, centered compensation
redistributes each residual over retained entries of the same sign:
\begin{equation}
\widetilde C_{i,v}
=
C_{i,v}+R_i^+/k_i^+\;(v\in S_i^+),\qquad
C_{i,v}-R_i^-/k_i^-\;(v\in S_i^-),\qquad
0\;(v\notin S_i).
\label{eq:centered_compensation}
\end{equation}
\paragraph{Preserved structure and approximation error.}
Compensation restores signed sums $M_i$ and $-M_i$, hence
$\sum_v\widetilde C_{i,v}=0$ and
$\|\widetilde C_i\|_1=\|C_i\|_1$. For fixed supports $S_i^\pm$, the uniform shift is the Euclidean projection
of the retained coefficients onto these constraints on correction mass.
The remaining approximation reflects losing tail coordinates
and representing their aggregate mass on the support. These two effects
are separated exactly by
\begin{equation}
\|\widetilde C_i-C_i\|_2^2=
\underbrace{\sum_{v\notin S_i}C_{i,v}^2}_{\text{support truncation}}
+\underbrace{\frac{(R_i^+)^2}{k_i^+}
+\frac{(R_i^-)^2}{k_i^-}}_{\text{residual compression}}.
\label{eq:compression_error}
\end{equation}
Empty sign-group terms are zero. Top-$C$ retains the coordinates contributing most to the first term.
Additional support reduces the second term by retaining more correction mass
and spreading the residual across more entries. The decomposition explains
why selection and compensation are complementary and why positions
carrying larger correction mass can benefit from more capacity.

\subsection{Sparse Backpropagation}
\label{sec:sparse_realization}

The selected support and compensated coefficients form a fixed sparse
plan $(S_i,\widetilde C_i)$ for backpropagation. Let
$h_i\in\mathbb R^{d_h}$ be the student hidden state and
$W\in\mathbb R^{|\mathcal V|\times d_h}$ the output projection.
The planning pass materializes $z_i=Wh_i$ transiently under no gradient
to construct the full-vocabulary statistics. These vocabulary-sized
tensors are discarded once the sparse plan is fixed.

The differentiable pass reconstructs only
$z_{i,S_i}=W_{S_i}h_i$, using the selected rows
$W_{S_i}\in\mathbb R^{|S_i|\times d_h}$.
To realize the planned coefficients without differentiating through their
construction, we use a linear surrogate with stop-gradient
$\operatorname{sg}(\cdot)$:
\begin{equation}
\widetilde{\mathcal L}
=\frac1N\sum_{i=1}^N\sum_{v\in S_i}
\operatorname{sg}(\widetilde C_{i,v})z_{i,v},
\qquad
\nabla_\theta\widetilde{\mathcal L}
=\frac1N\sum_{i=1}^N\sum_{v\in S_i}
\widetilde C_{i,v}\nabla_\theta z_{i,v}.
\label{eq:sparse_realization}
\end{equation}
Equation~(\ref{eq:sparse_realization}) realizes the prescribed sparse
coefficients exactly; the approximation is fixed by compression.
Full-vocabulary statistics inform this update, while autograd stores
vocabulary-axis intermediates only for the selected entries.

\paragraph{Distributed implementation.}
Allocation covers all valid positions in one optimizer update, collecting
position-level masses across gradient-accumulation microbatches and
data-parallel workers. Once the budgets are fixed, support
construction and selected-head backpropagation remain local.

\section{Experiments}
\label{sec:experiment}
\suppressfloats[t]

\subsection{Experimental Setup}
\label{sec:experimental_setup}

\paragraph{Datasets.}
We train on DAPO14k~\citep{guanning_dapo14k} for mathematics, the Chemistry
L3 multiple-choice partition of SciKnowEval V2~\citep{feng2024sciknoweval}
for chemistry, and MMR1-Math-RL-Data-v0~\citep{leng2025mmr1v0} for multimodal reasoning.
Mathematics evaluation covers MATH500~\citep{lightman2023verify},
Minerva~\citep{lewkowycz2022minerva}, AMC23~\citep{mathai2025amc23},
and AIME25~\citep{zhang2025aime25}.
Chemistry evaluation uses the held-out SciKnowEval questions and the
chemistry subset of GPQA-Diamond~\citep{rein2023gpqa}.
Multimodal evaluation covers MathVision~\citep{wang2024mathvision},
MathVista~\citep{lu2024mathvista}, We-Math~\citep{qiao2024wemath},
and MathVerse~\citep{zhang2024mathverse}.

\paragraph{Implementation.}
For mathematics and chemistry, we distill Qwen3-1.7B/4B into their Base
counterparts~\citep{yang2025qwen3} and report Mean@8 accuracy over eight
decoding runs. For multimodal reasoning, MMFineReason-2B/4B~\citep{lin2026mmfinereason}
teach Qwen3-VL-2B/4B-Instruct~\citep{bai2025qwen3vl}, evaluated by
single-generation accuracy with LMMs-Eval~\citep{zhang2024lmmseval}.
Further details are in Appendix~\ref{app:experimental_details}.

\ifdefined\fuseResultsStyleLoaded\else
\def\fuseResultsStyleLoaded{1}
\providecolor{nicegreen}{HTML}{23744D}
\providecolor{nicered}{HTML}{AA4B48}
\providecolor{fuseours}{HTML}{EFE8F8}
\providecolor{fusegroup}{HTML}{F0F1F3}
\providecolor{fuseavg}{HTML}{F8F5FC}
\providecommand{\posdelta}[1]{\makebox[1.50em][l]{\hspace{0.05em}\textsuperscript{\normalfont\fontsize{6}{6}\selectfont\textcolor{nicegreen}{+#1}}}}
\providecommand{\negdelta}[1]{\makebox[1.50em][l]{\hspace{0.05em}\textsuperscript{\normalfont\fontsize{6}{6}\selectfont\textcolor{nicered}{\ensuremath{-}#1}}}}
\providecommand{\nodelta}{\makebox[1.50em][l]{}}

\ifdefined\fuseresultstable\else
\newsavebox{\fusetablebox}
\newcommand{\fuseresultstable}[3]{%
  \begingroup
  \setlength{\tabcolsep}{0pt}%
  \sbox{\fusetablebox}{\begin{tabular}{#1}#3\end{tabular}}%
  \setlength{\tabcolsep}{\dimexpr(\linewidth-\wd\fusetablebox)/#2\relax}%
  \ifdim\tabcolsep<0pt\PackageError{fuse-tables}{Table exceeds text width}{Reduce column content before typesetting.}\fi
  \begin{tabular}{#1}#3\end{tabular}%
  \endgroup
}
\fi
\fi

\begin{table}[tb!]
    \centering
    \setlength{\abovecaptionskip}{6pt}
    \setlength{\belowcaptionskip}{4pt}
    \caption{\textbf{Mathematical reasoning and chemistry QA.} Superscripts show percentage-point changes from Base; bold indicates the best result.}
    \label{tab:main_results}
    \begin{minipage}{0.97\linewidth}
    \fontsize{9}{11}\selectfont
    \setlength{\tabcolsep}{1.5pt}
    \renewcommand{\arraystretch}{1.06}
    \fuseresultstable{l r r r r >{\columncolor{fuseavg}}r r r >{\columncolor{fuseavg}}r}{18}{
        \toprule
        \multirow{2}{*}{\textbf{Method}}
        & \multicolumn{5}{c}{\textbf{Mathematical Reasoning}}
        & \multicolumn{3}{c}{\textbf{Chemistry QA}} \\
        \cmidrule(lr){2-6} \cmidrule(lr){7-9}
        & \multicolumn{1}{c}{MATH} & \multicolumn{1}{c}{Minerva}
        & \multicolumn{1}{c}{AMC23} & \multicolumn{1}{c}{AIME25}
        & \multicolumn{1}{c}{\textbf{Avg.}} & \multicolumn{1}{c}{Chemistry}
        & \multicolumn{1}{c}{GPQA} & \multicolumn{1}{c}{\textbf{Avg.}} \\
        \midrule
        \rowcolor{fusegroup}
        \multicolumn{9}{l}{\textit{Qwen3-1.7B $\rightarrow$ Qwen3-1.7B-Base}} \\
        \addlinespace[1.5pt]
        Teacher
        & 72.3\nodelta & 28.8\nodelta & 44.4\nodelta & 8.3\nodelta & 38.4\nodelta
        & 42.0\nodelta & 26.5\nodelta & 34.2\nodelta \\
        Base
        & 46.5\nodelta & 15.3\nodelta & 26.9\nodelta & 2.9\nodelta & 22.9\nodelta
        & 24.9\nodelta & 22.9\nodelta & 23.9\nodelta \\
        \addlinespace[1.5pt]
        Sampled Token
        & 54.1\posdelta{7.6} & 18.2\posdelta{2.9} & 32.8\posdelta{5.9} & 3.8\posdelta{0.9} & 27.2\posdelta{4.3}
        & 34.3\posdelta{9.4} & 28.6\posdelta{5.7} & 31.5\posdelta{7.6} \\
        TopK
        & 63.6\posdelta{17.1} & 25.7\posdelta{10.4} & 34.1\posdelta{7.2} & 4.2\posdelta{1.3} & 31.9\posdelta{9.0}
        & 40.7\posdelta{15.8} & 29.2\posdelta{6.3} & 34.9\posdelta{11.0} \\
        \rowcolor{fuseours}
        \textbf{SparseOPD(Ours)}
        & \textbf{64.5}\posdelta{18.0} & \textbf{26.1}\posdelta{10.8} & \textbf{35.9}\posdelta{9.0} & \textbf{5.8}\posdelta{2.9} & \textbf{33.1}\posdelta{10.2}
        & \textbf{41.1}\posdelta{16.2} & \textbf{30.5}\posdelta{7.6} & \textbf{35.8}\posdelta{11.9} \\
        \addlinespace[1.5pt]
        \textit{Full Vocabulary}
        & \textit{63.4}\posdelta{16.9} & \textit{26.5}\posdelta{11.2} & \textit{35.3}\posdelta{8.4} & \textit{6.3}\posdelta{3.4} & \textit{32.9}\posdelta{10.0}
        & \textit{42.9}\posdelta{18.0} & \textit{31.1}\posdelta{8.2} & \textit{37.0}\posdelta{13.1} \\
        \midrule
        \rowcolor{fusegroup}
        \multicolumn{9}{l}{\textit{Qwen3-4B $\rightarrow$ Qwen3-4B-Base}} \\
        \addlinespace[1.5pt]
        Teacher
        & 83.3\nodelta & 38.8\nodelta & 64.1\nodelta & 19.6\nodelta & 51.4\nodelta
        & 41.5\nodelta & 31.9\nodelta & 36.7\nodelta \\
        Base
        & 56.8\nodelta & 18.7\nodelta & 41.6\nodelta & 8.3\nodelta & 31.3\nodelta
        & 34.3\nodelta & 29.7\nodelta & 32.0\nodelta \\
        \addlinespace[1.5pt]
        Sampled Token
        & 70.6\posdelta{13.8} & 29.7\posdelta{11.0} & 46.3\posdelta{4.7} & 12.1\posdelta{3.8} & 39.7\posdelta{8.4}
        & 38.0\posdelta{3.7} & 31.3\posdelta{1.6} & 34.7\posdelta{2.7} \\
        TopK
        & \textbf{77.4}\posdelta{20.6} & \textbf{30.1}\posdelta{11.4} & 45.9\posdelta{4.3} & 11.7\posdelta{3.4} & 41.3\posdelta{10.0}
        & 38.8\posdelta{4.5} & 29.8\posdelta{0.1} & 34.3\posdelta{2.3} \\
        \rowcolor{fuseours}
        \textbf{SparseOPD(Ours)}
        & 76.9\posdelta{20.1} & \textbf{30.1}\posdelta{11.4} & \textbf{50.0}\posdelta{8.4} & \textbf{15.8}\posdelta{7.5} & \textbf{43.2}\posdelta{11.9}
        & \textbf{39.7}\posdelta{5.4} & \textbf{33.2}\posdelta{3.5} & \textbf{36.5}\posdelta{4.5} \\
        \addlinespace[1.5pt]
        \textit{Full Vocabulary}
        & \textit{77.8}\posdelta{21.0} & \textit{29.8}\posdelta{11.1} & \textit{46.3}\posdelta{4.7} & \textit{12.5}\posdelta{4.2} & \textit{41.6}\posdelta{10.3}
        & \textit{40.7}\posdelta{6.4} & \textit{32.3}\posdelta{2.6} & \textit{36.5}\posdelta{4.5} \\
        \bottomrule
    }
    \end{minipage}
\end{table}

\begin{table}[tb!]
    \centering
    \setlength{\abovecaptionskip}{6pt}
    \setlength{\belowcaptionskip}{4pt}
    \caption{\textbf{Multimodal mathematical reasoning.} Superscripts show percentage-point changes from Student; bold indicates the best result.}
    \label{tab:mm_results}
    \begin{minipage}{0.97\linewidth}
    \fontsize{9}{11}\selectfont
    \setlength{\tabcolsep}{1.5pt}
    \renewcommand{\arraystretch}{1.06}
    \fuseresultstable{l r r r r >{\columncolor{fuseavg}}r}{12}{
        \toprule
        \textbf{Method}
        & \multicolumn{1}{c}{MathVision}
        & \multicolumn{1}{c}{MathVista}
        & \multicolumn{1}{c}{WeMath}
        & \multicolumn{1}{c}{MathVerse}
        & \multicolumn{1}{c}{\textbf{Avg.}} \\
        \midrule
        \rowcolor{fusegroup}
        \multicolumn{6}{l}{
        \textit{MMFineReason-2B $\rightarrow$ Qwen3-VL-2B-Instruct}
        } \\
        \addlinespace[1.5pt]
        Teacher
        & 24.4\nodelta
        & 65.1\nodelta
        & 66.1\nodelta
        & 46.0\nodelta
        & 50.4\nodelta \\
        Student
        & 15.9\nodelta
        & 59.3\nodelta
        & 56.2\nodelta
        & 36.5\nodelta
        & 42.0\nodelta \\
        \addlinespace[1.5pt]
        Sampled Token
        & 22.2\posdelta{6.3}
        & 60.8\posdelta{1.5}
        & 62.9\posdelta{6.7}
        & 40.7\posdelta{4.2}
        & 46.7\posdelta{4.7} \\
        TopK
        & 23.9\posdelta{8.1}
        & 62.2\posdelta{2.9}
        & 62.2\posdelta{6.0}
        & \textbf{43.3}\posdelta{6.7}
        & 47.9\posdelta{5.9} \\
        \rowcolor{fuseours}
        \textbf{SparseOPD(Ours)}
        & \textbf{24.2}\posdelta{8.3}
        & \textbf{63.8}\posdelta{4.5}
        & \textbf{64.8}\posdelta{8.6}
        & 42.0\posdelta{5.5}
        & \textbf{48.7}\posdelta{6.7} \\
        \addlinespace[1.5pt]
        \textit{Full Vocabulary}
        & \textit{24.9}\posdelta{9.0}
        & \textit{62.3}\posdelta{3.0}
        & \textit{63.2}\posdelta{7.0}
        & \textit{39.2}\posdelta{2.7}
        & \textit{47.4}\posdelta{5.4} \\
        \midrule
        \rowcolor{fusegroup}
        \multicolumn{6}{l}{
        \textit{MMFineReason-4B $\rightarrow$ Qwen3-VL-4B-Instruct}
        } \\
        \addlinespace[1.5pt]
        Teacher
        & 38.3\nodelta
        & 75.3\nodelta
        & 74.2\nodelta
        & 51.4\nodelta
        & 59.8\nodelta \\
        Student
        & 27.3\nodelta
        & 67.8\nodelta
        & 60.8\nodelta
        & 41.1\nodelta
        & 49.3\nodelta \\
        \addlinespace[1.5pt]
        Sampled Token
        & 30.5\posdelta{3.2}
        & 70.5\posdelta{2.7}
        & 68.5\posdelta{7.7}
        & 45.5\posdelta{4.4}
        & 53.8\posdelta{4.5} \\
        TopK
        & 35.7\posdelta{8.4}
        & 71.3\posdelta{3.5}
        & \textbf{75.5}\posdelta{14.7}
        & 54.1\posdelta{13.0}
        & 59.2\posdelta{9.9} \\
        \rowcolor{fuseours}
        \textbf{SparseOPD(Ours)}
        & \textbf{35.8}\posdelta{8.5}
        & \textbf{72.9}\posdelta{5.1}
        & \textbf{75.5}\posdelta{14.7}
        & \textbf{54.8}\posdelta{13.7}
        & \textbf{59.7}\posdelta{10.5} \\
        \addlinespace[1.5pt]
        \textit{Full Vocabulary}
        & \textit{36.1}\posdelta{8.8}
        & \textit{73.0}\posdelta{5.2}
        & \textit{75.0}\posdelta{14.2}
        & \textit{54.4}\posdelta{13.3}
        & \textit{59.6}\posdelta{10.4} \\
        \bottomrule
    }
    \end{minipage}
\end{table}

\subsection{Main Results}
\label{sec:main_results}

Tables~\ref{tab:main_results} and~\ref{tab:mm_results} compare SparseOPD
with Full Vocabulary, Sampled Token, and TopK across tasks and model scales.

\paragraph{Research Question 1: Can sparse execution preserve the training benefit of full-vocabulary correction?}
SparseOPD matches or exceeds the Full Vocabulary average in five of the six
task--scale settings at the reported precision. On Qwen3-4B mathematics, it achieves
43.2 compared with 41.6 for Full Vocabulary; both multimodal student
scales also exceed the dense reference. The exception is 1.7B chemistry,
where SparseOPD scores 35.8 against 37.0.
This pattern supports the separation underlying SparseOPD: much of the
training benefit of full-vocabulary correction survives when its correction
is compressed before differentiation.

\begin{figure}[t]
  \centering
  \graphicspath{{figures/}{iclr-2027-style-files/iclr2027/figures/}}
  \includegraphics[width=\linewidth]{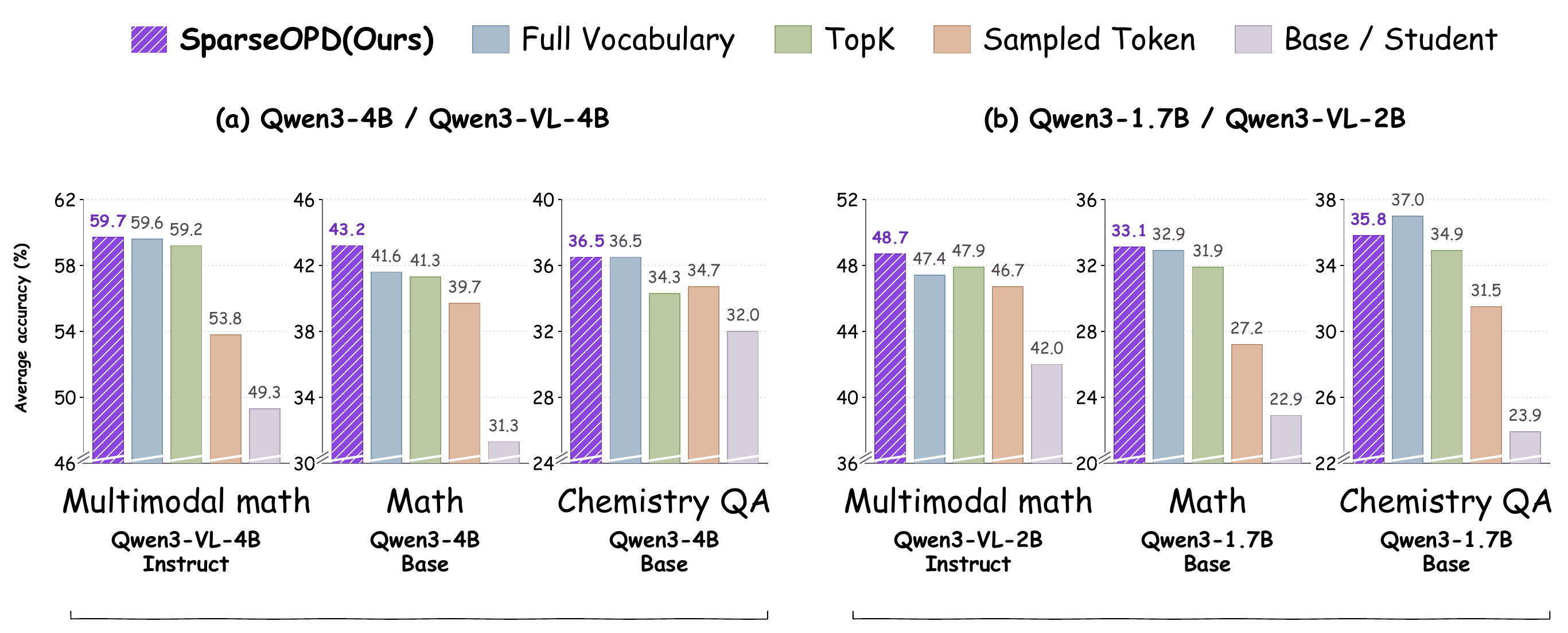}
  \setlength{\abovecaptionskip}{5pt}
  \setlength{\belowcaptionskip}{0pt}
    \vspace{-14pt}
\caption{\textbf{Task-average accuracy across student scales.}
SparseOPD outperforms both sparse baselines in all six settings and
matches or exceeds Full Vocabulary in five.}

  \label{fig:intro_results}
\end{figure}

\paragraph{Research Question 2: Does correction-aware compression make better use of the same sparse budget?}
With an average budget of 20 output-head coordinates per valid position,
SparseOPD outperforms TopK on all six task averages by 0.58--2.2 percentage
points. It also improves over Sampled Token by 1.8--5.97 points.
The comparison with TopK holds support size fixed while changing
how sparse updates are formed. Its consistent advantage supports using
the full-vocabulary correction to guide compression, making the same
differentiable support budget more effective.

\paragraph{Research Question 3: Do the task-average gains reflect broad improvements across benchmarks?}
The advantage over TopK extends to 16 of the 20 benchmark--scale
comparisons, with two ties at the reported precision. Gains include AMC23
and AIME25 on Qwen3-4B, as well as We-Math on Qwen3-VL-2B.
The remaining trade-offs occur on 4B MATH500 and 2B MathVerse, where TopK
retains a lead. Task-average improvements therefore reflect broad, but
not uniform, gains across individual benchmarks.

\subsection{Ablation Study}
\label{sec:ablation}

\paragraph{Research Question 4: Do selection, compensation, and global allocation each contribute under the same sparse budget?}
Table~\ref{tab:ablation_results} tests the role of selection, compensation,
and allocation in preserving useful corrections. Each variant changes
one component on Qwen3-4B mathematical reasoning, holding the remaining
components and total support budget $KN$ fixed, with $K=20$.

\paragraph{Signed Top-$C$ selection.}
The largest ablation drop comes from replacing coefficient-magnitude
ranking with student-probability ranking within each sign group: average
accuracy falls by 2.0 percentage points. Student likelihood alone does
not capture the teacher--student discrepancy that drives correction
strength. The result supports retaining coordinates according to their
contribution to the full-vocabulary update.

\paragraph{Centered compensation.}
Retaining strong coefficients does not recover all useful
correction. Using their original values without compensation lowers average
accuracy by 1.4 percentage points. Compensation preserves the aggregate
promoting and suppressing signal carried by discarded candidates,
allowing the retained support to represent more than its selected
coefficients alone.

\paragraph{Batch-global allocation.}
The support budget is more effective when it can follow correction demand
across positions. Replacing global allocation with a fixed budget of $K$
coordinates per valid position lowers average accuracy by 0.9 percentage
points. Since total support is unchanged, this comparison supports
redistributing capacity toward positions with stronger corrections.

\begin{table}[tb!]
    \centering
    \small
    \setlength{\abovecaptionskip}{8pt}
    \setlength{\belowcaptionskip}{5pt}
\caption{\textbf{Ablation of SparseOPD on mathematical reasoning.}
Results use Qwen3-4B $\rightarrow$ Qwen3-4B-Base under the same support
budget. Red superscripts show drops
from SparseOPD.}
    \label{tab:ablation_results}

    \setlength{\tabcolsep}{8pt}
    \renewcommand{\arraystretch}{1.08}

    \begin{tabular}{lrrrr>{\columncolor{fuseavg}}r}
        \toprule
        \textbf{Variant}
        & \multicolumn{1}{c}{MATH500}
        & \multicolumn{1}{c}{Minerva}
        & \multicolumn{1}{c}{AMC23}
        & \multicolumn{1}{c}{AIME25}
        & \multicolumn{1}{c}{\textbf{Avg.}} \\
        \midrule

        \rowcolor{fuseours}
        \textbf{SparseOPD(Ours)}
        & \textbf{76.9}\nodelta
        & \textbf{30.1}\nodelta
        & \textbf{50.0}\nodelta
        & \textbf{15.8}\nodelta
        & \textbf{43.2}\nodelta \\

        Student Probability Ranking
        & 76.0\negdelta{0.9}
        & 29.0\negdelta{1.1}
        & 45.6\negdelta{4.4}
        & 14.2\negdelta{1.6}
        & 41.2\negdelta{2.0} \\

        w/o Centered Compensation
        & 76.3\negdelta{0.6}
        & 29.2\negdelta{0.9}
        & 47.5\negdelta{2.5}
        & 14.2\negdelta{1.6}
        & 41.8\negdelta{1.4} \\

        w/o Batch-global Allocation
        & 76.6\negdelta{0.3}
        & 29.6\negdelta{0.5}
        & 48.1\negdelta{1.9}
        & 15.0\negdelta{0.8}
        & 42.3\negdelta{0.9} \\

        \bottomrule
    \end{tabular}
\vspace{-8pt}
\end{table}

\subsection{Analysis and Discussion}
\label{sec:analysis_discussion}

Gradient fidelity tests preservation of the reference update, while
efficiency profiling tests the cost of realizing it. Both analyses use
Qwen3-4B as the teacher and Qwen3-4B-Base as the student, with mathematical
reasoning batches for the gradient comparison.

\paragraph{Research Question 5: Does SparseOPD preserve the full-vocabulary update?}
At matched student states and on-policy batches, let $g$ and $g_F$ denote
the candidate and Full Vocabulary parameter gradients before clipping.
We report cosine similarity, relative $\ell_2$ error
$\|g-g_F\|_2/\|g_F\|_2$, norm ratio $\|g\|_2/\|g_F\|_2$, and sign agreement,
excluding jointly zero coordinates and treating other zeros as nonnegative.
Metrics are computed at each recorded step and averaged over full-parameter training.

\begin{wraptable}{r}{0.53\linewidth}
  \vspace{-16pt}
  \centering
  \small
  \setlength{\abovecaptionskip}{3pt}
  \setlength{\belowcaptionskip}{4pt}
  \caption{\textbf{Training-averaged gradient fidelity to Full Vocabulary OPD.}
Bold marks the best.}
\label{tab:gradient_fidelity}

  \fontsize{8}{10}\selectfont
  \renewcommand{\arraystretch}{1.04}
  \fuseresultstable{ccccc}{10}{
    \toprule
    Method & Cos. $\uparrow$ & Rel. $\ell_2\downarrow$
    & Norm $\to1$ & Sign $\uparrow$ \\
    \midrule
    Full Vocabulary & 1.00 & 0.00 & 1.00 & 1.00 \\
    \midrule
    Sampled Token & 0.29 & 5.43 & 5.63 & 0.59 \\
    TopK & 0.90 & 0.45 & 1.05 & 0.83 \\
    \rowcolor[HTML]{EFE8F8}
    \textbf{SparseOPD(Ours)} & \textbf{0.99} & \textbf{0.15}
    & \textbf{0.99} & \textbf{0.91} \\
    \bottomrule
  }
\end{wraptable}

SparseOPD is closest to the reference on all four metrics
(Table~\ref{tab:gradient_fidelity}). Its relative $\ell_2$ error is
approximately one third of TopK's (0.15 versus 0.45), while cosine
similarity and norm ratio are both near one. SparseOPD also achieves the
highest sign agreement, supporting preservation of the dense update
through sparse execution.
\WFclear
\par\vspace{4pt}

\paragraph{Research Question 6: How much backward memory does sparse execution save?}
Decoupling correction construction from differentiation reduces memory
(Table~\ref{tab:execution_memory}).

\begin{wraptable}{r}{0.53\linewidth}
  \vspace{-16pt}
  \centering
  \small
  \setlength{\abovecaptionskip}{3pt}
  \setlength{\belowcaptionskip}{4pt}
  \caption{\textbf{Memory usage.}
  Peak extra GPU allocation above the pre-step baseline
  (Appendix~\ref{app:efficiency_measurements}).}
  \label{tab:execution_memory}

  \newcommand{\effcell}[2]{\textbf{#1}\,\textcolor{nicegreen}{%
    {\fontsize{7.5}{8.5}\selectfont(\ensuremath{\downarrow}#2\%)}}}

  \fontsize{8}{10}\selectfont
  \renewcommand{\arraystretch}{1.04}
  \fuseresultstable{cccc}{9}{
    \toprule
    \multirow{2}{*}{Length}
    & \multirow{2}{*}{Method}
    & \multicolumn{2}{c}{Memory (GiB)} \\
    \cmidrule(lr){3-4}
    & & Forward & Backward \\
    \midrule
    4K & Full Vocabulary & 12.82 & 10.09 \\
    \rowcolor{fuseours}
       & \textbf{SparseOPD(Ours)}
       & \effcell{1.94}{84.9} & \effcell{3.59}{64.5} \\
    \midrule
    8K & Full Vocabulary & 25.50 & 20.05 \\
    \rowcolor{fuseours}
       & \textbf{SparseOPD(Ours)}
       & \effcell{3.88}{84.8} & \effcell{5.91}{70.5} \\
    \midrule
    16K & Full Vocabulary & 50.85 & 39.96 \\
    \rowcolor{fuseours}
       & \textbf{SparseOPD(Ours)}
       & \effcell{7.76}{84.7} & \effcell{10.54}{73.6} \\
    \bottomrule
  }
\end{wraptable}
Full-vocabulary statistics are computed without an autograd graph and
discarded after planning; backpropagation uses only selected output-head
rows. This reduces backward memory by 70.5\% at 8K, with larger savings
at longer sequences. Backward time improves more modestly because the
Transformer backbone still requires full backpropagation, while additional
planning adds overhead (Appendix~\ref{app:efficiency_measurements}).
\WFclear

\section{Related Work}
On-policy distillation trains on student-generated trajectories with teacher
feedback~\citep{agarwal2024onpolicy,ko2024distillm,ye2025gad}.
For large language models, reverse-KL objectives provide full-vocabulary
supervision~\citep{gu2024minillm,zhong2024revisiting}.
Prior work reduces this cost by sampling vocabulary entries or restricting
supervision to a small support~\citep{anshumann2025sparse,huang2026tailaware}.
SparseOPD instead forms the full-vocabulary correction first and compresses it
for sparse backpropagation. See Appendix~\ref{sec:related_work} for details.

\section{Conclusions}
\label{sec:conclusion}
We introduced SparseOPD, an on-policy distillation method that separates the construction of full-vocabulary corrections from sparse backpropagation. By selecting entries according to correction magnitude and compensating for discarded correction mass, SparseOPD compresses the update while retaining information from the complete teacher and student distributions. Across mathematical reasoning, chemistry question answering, and multimodal reasoning, our experiments show that sparse execution can preserve much of the training benefit of full-vocabulary supervision. Complementary analyses demonstrate close agreement with the full-vocabulary gradient and substantially lower peak backward memory in the measured setting. Overall, SparseOPD preserves full-vocabulary correction while substantially reducing the memory required for backpropagation.




\subsection*{Reproducibility statement}
SparseOPD and the comparison baselines are specified in
Sections~\ref{sec:rethinking} and~\ref{sec:method}; the datasets,
model configurations, evaluation metrics, and benchmark-level results
are reported in Section~\ref{sec:experiment}. Training splits,
optimization and distillation settings, evaluation and decoding
protocols, and efficiency-measurement procedures are provided in
Appendix~\ref{app:experimental_details}.

\bibliography{iclr2027_conference}

@inproceedings{agarwal2024onpolicy,
  title={On-Policy Distillation of Language Models: Learning from Self-Generated Mistakes},
  author={Rishabh Agarwal and Nino Vieillard and Yongchao Zhou and Piotr Stanczyk and Sabela Ramos Garea and Matthieu Geist and Olivier Bachem},
  booktitle={The Twelfth International Conference on Learning Representations},
  year={2024},
  url={https://proceedings.iclr.cc/paper_files/paper/2024/hash/5be69a584901a26c521c2b51e40a4c20-Abstract-Conference.html}
}

@inproceedings{gu2024minillm,
  title={{MiniLLM}: Knowledge Distillation of Large Language Models},
  author={Yuxian Gu and Li Dong and Furu Wei and Minlie Huang},
  booktitle={The Twelfth International Conference on Learning Representations},
  year={2024},
  url={https://proceedings.iclr.cc/paper_files/paper/2024/hash/8ac015d409635f196f9e3e9dcfb9a94e-Abstract-Conference.html}
}

@misc{yang2025qwen3,
  title={{Qwen3} Technical Report},
  author={An Yang and Anfeng Li and Baosong Yang and Beichen Zhang and Binyuan Hui and Bo Zheng and Bowen Yu and Chang Gao and Chengen Huang and Chenxu Lv and Chujie Zheng and Dayiheng Liu and Fan Zhou and Fei Huang and Feng Hu and Hao Ge and Haoran Wei and Huan Lin and Jialong Tang and Jian Yang and Jianhong Tu and Jianwei Zhang and Jianxin Yang and Jiaxi Yang and Jing Zhou and Jingren Zhou and Junyang Lin and Kai Dang and Keqin Bao and Kexin Yang and Le Yu and Lianghao Deng and Mei Li and Mingfeng Xue and Mingze Li and Pei Zhang and Peng Wang and Qin Zhu and Rui Men and Ruize Gao and Shixuan Liu and Shuang Luo and Tianhao Li and Tianyi Tang and Wenbiao Yin and Xingzhang Ren and Xinyu Wang and Xinyu Zhang and Xuancheng Ren and Yang Fan and Yang Su and Yichang Zhang and Yinger Zhang and Yu Wan and Yuqiong Liu and Zekun Wang and Zeyu Cui and Zhenru Zhang and Zhipeng Zhou and Zihan Qiu},
  year={2025},
  howpublished={arXiv preprint arXiv:2505.09388},
  url={https://arxiv.org/abs/2505.09388}
}

@misc{bai2025qwen3vl,
  title={{Qwen3-VL} Technical Report},
  author={Shuai Bai and Yuxuan Cai and Ruizhe Chen and Keqin Chen and Xionghui Chen and Zesen Cheng and Lianghao Deng and Wei Ding and Chang Gao and Chunjiang Ge and Wenbin Ge and Zhifang Guo and Qidong Huang and Jie Huang and Fei Huang and Binyuan Hui and Shutong Jiang and Zhaohai Li and Mingsheng Li and Mei Li and Kaixin Li and Zicheng Lin and Junyang Lin and Xuejing Liu and Jiawei Liu and Chenglong Liu and Yang Liu and Dayiheng Liu and Shixuan Liu and Dunjie Lu and Ruilin Luo and Chenxu Lv and Rui Men and Lingchen Meng and Xuancheng Ren and Xingzhang Ren and Sibo Song and Yuchong Sun and Jun Tang and Jianhong Tu and Jianqiang Wan and Peng Wang and Pengfei Wang and Qiuyue Wang and Yuxuan Wang and Tianbao Xie and Yiheng Xu and Haiyang Xu and Jin Xu and Zhibo Yang and Mingkun Yang and Jianxin Yang and An Yang and Bowen Yu and Fei Zhang and Hang Zhang and Xi Zhang and Bo Zheng and Humen Zhong and Jingren Zhou and Fan Zhou and Jing Zhou and Yuanzhi Zhu and Ke Zhu},
  year={2025},
  howpublished={arXiv preprint arXiv:2511.21631},
  url={https://arxiv.org/abs/2511.21631}
}

@misc{lin2026mmfinereason,
  title={{MMFineReason}: Closing the Multimodal Reasoning Gap via Open Data-Centric Methods},
  author={Honglin Lin and Zheng Liu and Yun Zhu and Chonghan Qin and Juekai Lin and Xiaoran Shang and Conghui He and Wentao Zhang and Lijun Wu},
  year={2026},
  howpublished={arXiv preprint arXiv:2601.21821},
  url={https://arxiv.org/abs/2601.21821}
}

@misc{guanning_dapo14k,
  title={{DAPO14k}},
  author={{Guanning-AI}},
  year={n.d.},
  howpublished={Hugging Face dataset},
  url={https://huggingface.co/datasets/guanning-ai/dapo14k},
  note={Accessed September 9, 2026}
}

@misc{feng2024sciknoweval,
  title={{SciKnowEval}: Evaluating Multi-level Scientific Knowledge of Large Language Models},
  author={Kehua Feng and Xinyi Shen and Weijie Wang and Xiang Zhuang and Yuqi Tang and Qiang Zhang and Keyan Ding},
  year={2024},
  howpublished={arXiv preprint arXiv:2406.09098, version 4},
  url={https://arxiv.org/abs/2406.09098v4},
  note={Version 4 revised in 2025; SciKnowEval V2 dataset: \url{https://huggingface.co/datasets/hicai-zju/SciKnowEval}}
}

@misc{leng2025mmr1v0,
  title={{MMR1}: Advancing the Frontiers of Multimodal Reasoning},
  author={Sicong Leng and Jing Wang and Jiaxi Li and Hao Zhang and Zhiqiang Hu and Boqiang Zhang and Hang Zhang and Yuming Jiang and Xin Li and Deli Zhao and Fan Wang and Yu Rong and Aixin Sun and Shijian Lu},
  year={2025},
  howpublished={Open-source project, v0 release},
  url={https://github.com/LengSicong/MMR1/tree/mmr1_v0},
  note={Dataset: \url{https://huggingface.co/datasets/MMR1/MMR1-Math-RL-Data-v0}}
}

@misc{lightman2023verify,
  title={Let's Verify Step by Step},
  author={Hunter Lightman and Vineet Kosaraju and Yura Burda and Harri Edwards and Bowen Baker and Teddy Lee and Jan Leike and John Schulman and Ilya Sutskever and Karl Cobbe},
  year={2023},
  howpublished={arXiv preprint arXiv:2305.20050},
  url={https://arxiv.org/abs/2305.20050}
}

@misc{lewkowycz2022minerva,
  title={Solving Quantitative Reasoning Problems with Language Models},
  author={Aitor Lewkowycz and Anders Andreassen and David Dohan and Ethan Dyer and Henryk Michalewski and Vinay Ramasesh and Ambrose Slone and Cem Anil and Imanol Schlag and Theo Gutman-Solo and Yuhuai Wu and Behnam Neyshabur and Guy Gur-Ari and Vedant Misra},
  year={2022},
  howpublished={arXiv preprint arXiv:2206.14858},
  url={https://arxiv.org/abs/2206.14858}
}

@misc{mathai2025amc23,
  title={{AMC23}: American Mathematics Competitions 2023},
  author={{Math-AI Team}},
  year={2025},
  howpublished={Hugging Face dataset},
  url={https://huggingface.co/datasets/math-ai/amc23}
}

@misc{zhang2025aime25,
  title={American Invitational Mathematics Examination ({AIME}) 2025},
  author={Yifan Zhang and {Math-AI Team}},
  year={2025},
  howpublished={Hugging Face dataset},
  url={https://huggingface.co/datasets/math-ai/aime25}
}

@misc{rein2023gpqa,
  title={{GPQA}: A Graduate-Level Google-Proof {Q\&A} Benchmark},
  author={David Rein and Betty Li Hou and Asa Cooper Stickland and Jackson Petty and Richard Yuanzhe Pang and Julien Dirani and Julian Michael and Samuel R. Bowman},
  year={2023},
  howpublished={arXiv preprint arXiv:2311.12022},
  url={https://arxiv.org/abs/2311.12022}
}

@misc{wang2024mathvision,
  title={Measuring Multimodal Mathematical Reasoning with {MATH-Vision} Dataset},
  author={Ke Wang and Junting Pan and Weikang Shi and Zimu Lu and Mingjie Zhan and Hongsheng Li},
  year={2024},
  howpublished={arXiv preprint arXiv:2402.14804},
  url={https://arxiv.org/abs/2402.14804}
}

@inproceedings{lu2024mathvista,
  title={{MathVista}: Evaluating Mathematical Reasoning of Foundation Models in Visual Contexts},
  author={Pan Lu and Hritik Bansal and Tony Xia and Jiacheng Liu and Chunyuan Li and Hannaneh Hajishirzi and Hao Cheng and Kai-Wei Chang and Michel Galley and Jianfeng Gao},
  booktitle={The Twelfth International Conference on Learning Representations},
  year={2024},
  url={https://proceedings.iclr.cc/paper_files/paper/2024/hash/663bce02a0050c4a11f1eb8a7f1429d3-Abstract-Conference.html}
}

@misc{qiao2024wemath,
  title={{We-Math}: Does Your Large Multimodal Model Achieve Human-like Mathematical Reasoning?},
  author={Runqi Qiao and Qiuna Tan and Guanting Dong and Minhui Wu and Chong Sun and Xiaoshuai Song and Zhuoma GongQue and Shanglin Lei and Zhe Wei and Miaoxuan Zhang and Runfeng Qiao and Yifan Zhang and Xiao Zong and Yida Xu and Muxi Diao and Zhimin Bao and Chen Li and Honggang Zhang},
  year={2024},
  howpublished={arXiv preprint arXiv:2407.01284},
  url={https://arxiv.org/abs/2407.01284}
}

@misc{zhang2024mathverse,
  title={{MathVerse}: Does Your Multi-modal {LLM} Truly See the Diagrams in Visual Math Problems?},
  author={Renrui Zhang and Dongzhi Jiang and Yichi Zhang and Haokun Lin and Ziyu Guo and Pengshuo Qiu and Aojun Zhou and Pan Lu and Kai-Wei Chang and Peng Gao and Hongsheng Li},
  year={2024},
  howpublished={arXiv preprint arXiv:2403.14624},
  url={https://arxiv.org/abs/2403.14624}
}

@misc{zhang2024lmmseval,
  title={{LMMs-Eval}: Reality Check on the Evaluation of Large Multimodal Models},
  author={Kaichen Zhang and Bo Li and Peiyuan Zhang and Fanyi Pu and Joshua Adrian Cahyono and Kairui Hu and Shuai Liu and Yuanhan Zhang and Jingkang Yang and Chunyuan Li and Ziwei Liu},
  year={2024},
  howpublished={arXiv preprint arXiv:2407.12772},
  url={https://arxiv.org/abs/2407.12772}
}

@misc{hu2021lora,
  title={{LoRA}: Low-Rank Adaptation of Large Language Models},
  author={Edward J. Hu and Yelong Shen and Phillip Wallis and Zeyuan Allen-Zhu and Yuanzhi Li and Shean Wang and Lu Wang and Weizhu Chen},
  year={2021},
  howpublished={arXiv preprint arXiv:2106.09685},
  url={https://arxiv.org/abs/2106.09685}
}

@inproceedings{loshchilov2019adamw,
  title={Decoupled Weight Decay Regularization},
  author={Ilya Loshchilov and Frank Hutter},
  booktitle={International Conference on Learning Representations},
  year={2019},
  url={https://arxiv.org/abs/1711.05101}
}

@misc{ye2026opcd,
  title        = {On-Policy Context Distillation for Language Models},
  author       = {Tianzhu Ye and Li Dong and Xun Wu and Shaohan Huang and Furu Wei},
  year         = {2026},
  eprint       = {2602.12275},
  archivePrefix = {arXiv},
  primaryClass = {cs.CL},
  url          = {https://arxiv.org/abs/2602.12275}
}

@misc{ren2026topd,
  title        = {Trace-Based On-Policy Distillation for Masked Diffusion Language Models},
  author       = {Haolin Ren and Ziyang Huang and Chenhao Yuan and Jun Zhao and Kang Liu},
  year         = {2026},
  eprint       = {2607.16872},
  archivePrefix = {arXiv},
  primaryClass = {cs.CL},
  url          = {https://arxiv.org/abs/2607.16872}
}

@inproceedings{anshumann2025sparse,
  title     = {Sparse Logit Sampling: Accelerating Knowledge Distillation in {LLM}s},
  author    = {Anshumann and Mohd Abbas Zaidi and Akhil Kedia and Jinwoo Ahn
               and Taehwak Kwon and Kangwook Lee and Haejun Lee and Joohyung Lee},
  booktitle = {Proceedings of the 63rd Annual Meeting of the Association for Computational Linguistics (Volume 1: Long Papers)},
  pages     = {18085--18108},
  year      = {2025},
  address   = {Vienna, Austria},
  publisher = {Association for Computational Linguistics},
  doi       = {10.18653/v1/2025.acl-long.885},
  url       = {https://aclanthology.org/2025.acl-long.885/}
}

@misc{huang2026tailaware,
  title        = {Tail-Aware Top-$k$ On-Policy Distillation},
  author       = {Huipeng Huang and Hongxin Wei},
  year         = {2026},
  eprint       = {2608.14728},
  archivePrefix = {arXiv},
  primaryClass = {cs.LG},
  url          = {https://arxiv.org/abs/2608.14728}
}

@inproceedings{ko2024distillm,
  title        = {{DistiLLM}: Towards Streamlined Distillation for Large Language Models},
  author       = {Jongwoo Ko and Sungnyun Kim and Tianyi Chen and Se-Young Yun},
  booktitle    = {Proceedings of the 41st International Conference on Machine Learning},
  pages        = {24872--24895},
  year         = {2024},
  publisher    = {PMLR},
  url          = {https://proceedings.mlr.press/v235/ko24c.html}
}

@inproceedings{zhong2024revisiting,
  title        = {Revisiting Knowledge Distillation for Autoregressive Language Models},
  author       = {Qihuang Zhong and Liang Ding and Li Shen and Juhua Liu and Bo Du and Dacheng Tao},
  booktitle    = {Proceedings of the 62nd Annual Meeting of the Association for Computational Linguistics (Volume 1: Long Papers)},
  pages        = {10900--10913},
  year         = {2024},
  publisher    = {Association for Computational Linguistics},
  doi          = {10.18653/v1/2024.acl-long.587},
  url          = {https://aclanthology.org/2024.acl-long.587/}
}

@inproceedings{kim2024promptkd,
  title        = {{PromptKD}: Distilling Student-Friendly Knowledge for Generative Language Models via Prompt Tuning},
  author       = {Gyeongman Kim and Doohyuk Jang and Eunho Yang},
  booktitle    = {Findings of the Association for Computational Linguistics: EMNLP 2024},
  pages        = {6266--6282},
  year         = {2024},
  publisher    = {Association for Computational Linguistics},
  doi          = {10.18653/v1/2024.findings-emnlp.364},
  url          = {https://aclanthology.org/2024.findings-emnlp.364/}
}

@inproceedings{zhang2024dualspace,
  title        = {Dual-Space Knowledge Distillation for Large Language Models},
  author       = {Songming Zhang and Xue Zhang and Zengkui Sun and Yufeng Chen and Jinan Xu},
  booktitle    = {Proceedings of the 2024 Conference on Empirical Methods in Natural Language Processing},
  pages        = {18164--18181},
  year         = {2024},
  publisher    = {Association for Computational Linguistics},
  doi          = {10.18653/v1/2024.emnlp-main.1010},
  url          = {https://aclanthology.org/2024.emnlp-main.1010/}
}

@inproceedings{lee2024mentorkd,
  title        = {{Mentor-KD}: Making Small Language Models Better Multi-step Reasoners},
  author       = {Hojae Lee and Junho Kim and SangKeun Lee},
  booktitle    = {Proceedings of the 2024 Conference on Empirical Methods in Natural Language Processing},
  pages        = {17643--17658},
  year         = {2024},
  publisher    = {Association for Computational Linguistics},
  doi          = {10.18653/v1/2024.emnlp-main.977},
  url          = {https://aclanthology.org/2024.emnlp-main.977/}
}

@inproceedings{zhou2024distillspec,
  title        = {{DistillSpec}: Improving Speculative Decoding via Knowledge Distillation},
  author       = {Yongchao Zhou and Kaifeng Lyu and Ankit Singh Rawat and Aditya Krishna Menon and Afshin Rostamizadeh and Sanjiv Kumar and Jean-Fran{\c{c}}ois Kagy and Rishabh Agarwal},
  booktitle    = {The Twelfth International Conference on Learning Representations},
  year         = {2024},
  url          = {https://proceedings.iclr.cc/paper_files/paper/2024/hash/8766fbc68e1ed1cdef712ce273e0a363-Abstract-Conference.html}
}

@misc{riviere2024gemma2,
  title        = {{Gemma 2}: Improving Open Language Models at a Practical Size},
  author       = {{Gemma Team}},
  year         = {2024},
  eprint       = {2408.00118},
  archivePrefix = {arXiv},
  primaryClass = {cs.CL},
  url          = {https://arxiv.org/abs/2408.00118}
}

@misc{boizard2024uld,
  title        = {Towards Cross-Tokenizer Distillation: the Universal Logit Distillation Loss for {LLM}s},
  author       = {Nicolas Boizard and Kevin El Haddad and C{\'e}line Hudelot and Pierre Colombo},
  year         = {2024},
  eprint       = {2402.12030},
  archivePrefix = {arXiv},
  primaryClass = {cs.CL},
  url          = {https://arxiv.org/abs/2402.12030}
}

@inproceedings{busbridge2025scaling,
  title        = {Distillation Scaling Laws},
  author       = {Dan Busbridge and Amitis Shidani and Floris Weers and Jason Ramapuram and Etai Littwin and Russell Webb},
  booktitle    = {Proceedings of the 42nd International Conference on Machine Learning},
  pages        = {5977--6045},
  year         = {2025},
  publisher    = {PMLR},
  url          = {https://proceedings.mlr.press/v267/busbridge25a.html}
}

@inproceedings{wu2025akl,
  title        = {Rethinking Kullback--Leibler Divergence in Knowledge Distillation for Large Language Models},
  author       = {Taiqiang Wu and Chaofan Tao and Jiahao Wang and Runming Yang and Zhe Zhao and Ngai Wong},
  booktitle    = {Proceedings of the 31st International Conference on Computational Linguistics},
  pages        = {5737--5755},
  year         = {2025},
  publisher    = {Association for Computational Linguistics},
  url          = {https://aclanthology.org/2025.coling-main.383/}
}

@inproceedings{zhang2025aligndistil,
  title        = {{AlignDistil}: Token-Level Language Model Alignment as Adaptive Policy Distillation},
  author       = {Songming Zhang and Xue Zhang and Tong Zhang and Bojie Hu and Yufeng Chen and Jinan Xu},
  booktitle    = {Proceedings of the 63rd Annual Meeting of the Association for Computational Linguistics (Volume 1: Long Papers)},
  pages        = {19791--19807},
  year         = {2025},
  publisher    = {Association for Computational Linguistics},
  doi          = {10.18653/v1/2025.acl-long.972},
  url          = {https://aclanthology.org/2025.acl-long.972/}
}

@misc{ye2025gad,
  title        = {Black-Box On-Policy Distillation of Large Language Models},
  author       = {Tianzhu Ye and Li Dong and Zewen Chi and Xun Wu and Shaohan Huang and Furu Wei},
  year         = {2025},
  eprint       = {2511.10643},
  archivePrefix = {arXiv},
  primaryClass = {cs.CL},
  url          = {https://arxiv.org/abs/2511.10643}
}

@misc{bousselham2025vold,
  title        = {{VOLD}: Reasoning Transfer from {LLM}s to Vision-Language Models via On-Policy Distillation},
  author       = {Walid Bousselham and Hilde Kuehne and Cordelia Schmid},
  year         = {2025},
  eprint       = {2510.23497},
  archivePrefix = {arXiv},
  primaryClass = {cs.CV},
  url          = {https://arxiv.org/abs/2510.23497}
}

@misc{fan2025ketchup,
  title        = {{KETCHUP}: K-Step Return Estimation for Sequential Knowledge Distillation},
  author       = {Jiabin Fan and Guoqing Luo and Michael Bowling and Lili Mou},
  year         = {2025},
  eprint       = {2504.19024},
  archivePrefix = {arXiv},
  primaryClass = {cs.CL},
  url          = {https://arxiv.org/abs/2504.19024}
}

@misc{xu2025kdrl,
  title        = {{KDRL}: Post-Training Reasoning {LLM}s via Unified Knowledge Distillation and Reinforcement Learning},
  author       = {Hongling Xu and Qi Zhu and Heyuan Deng and Jinpeng Li and Lu Hou and Yasheng Wang and Lifeng Shang and Ruifeng Xu and Fei Mi},
  year         = {2025},
  eprint       = {2506.02208},
  archivePrefix = {arXiv},
  primaryClass = {cs.LG},
  url          = {https://arxiv.org/abs/2506.02208}
}

@misc{xu2025rlkd,
  title        = {{RLKD}: Distilling {LLM}s' Reasoning via Reinforcement Learning},
  author       = {Shicheng Xu and Liang Pang and Yunchang Zhu and Jia Gu and Zihao Wei and Jingcheng Deng and Feiyang Pan and Huawei Shen and Xueqi Cheng},
  year         = {2025},
  eprint       = {2505.16142},
  archivePrefix = {arXiv},
  primaryClass = {cs.CL},
  url          = {https://arxiv.org/abs/2505.16142}
}

@misc{song2026survey,
  title        = {A Survey of On-Policy Distillation for Large Language Models},
  author       = {Mingyang Song and Mao Zheng},
  year         = {2026},
  eprint       = {2604.00626},
  archivePrefix = {arXiv},
  primaryClass = {cs.LG},
  url          = {https://arxiv.org/abs/2604.00626}
}

@misc{sun2026easyopd,
  title        = {{EasyOPD}: An Easy-to-use On-Policy Distillation Framework for Large Language Models},
  author       = {Jie Sun and Mao Zheng and Mingyang Song and Qiyong Zhong and Gengsheng Li and Zhepei Hong and Chang Wu and Pengfei Liu and Junfeng Fang and Xiang Wang},
  year         = {2026},
  eprint       = {2607.11012},
  archivePrefix = {arXiv},
  primaryClass = {cs.CL},
  url          = {https://arxiv.org/abs/2607.11012}
}

@misc{agrawal2026distil,
  title        = {Reinforcement Learning from Rich Feedback with Distributional {DAgger}},
  author       = {Rishabh Agrawal and Jacob Fein-Ashley and Paria Rashidinejad},
  year         = {2026},
  eprint       = {2606.05152},
  archivePrefix = {arXiv},
  primaryClass = {cs.LG},
  url          = {https://arxiv.org/abs/2606.05152}
}

@misc{armandpour2026unmasking,
  title        = {Unmasking On-Policy Distillation: Where It Helps, Where It Hurts, and Why},
  author       = {Mohammadreza Armandpour and Fatih Ilhan and David Harrison and Ajay Jaiswal and Duc N. M. Hoang and Fartash Faghri and Yizhe Zhang and Minsik Cho and Mehrdad Farajtabar},
  year         = {2026},
  eprint       = {2605.10889},
  archivePrefix = {arXiv},
  primaryClass = {cs.LG},
  url          = {https://arxiv.org/abs/2605.10889}
}

@misc{byeon2026tutor,
  title        = {Be My Tutor: On-Policy Co-Distillation for Mutual {LLM} Improvement via Peer Feedback},
  author       = {Woohyeon Byeon and Jiwon Jeon and Jeonghye Kim and Youngchul Sung},
  year         = {2026},
  eprint       = {2606.14368},
  archivePrefix = {arXiv},
  primaryClass = {cs.LG},
  url          = {https://arxiv.org/abs/2606.14368}
}

@misc{cai2026imagineopd,
  title        = {Thinking Without Images: Internalizing Visual Manipulation with On-Policy Self-Distillation},
  author       = {Yishuo Cai and Jiahui Liu and Yuanxin Liu and Haobo Deng and Linli Yao and Yuhao Zheng and Kun Ouyang and Zhimo Li and Ziyue Wang and Xu Sun and Haoli Bai and Xiaohui Li},
  year         = {2026},
  eprint       = {2606.08719},
  archivePrefix = {arXiv},
  primaryClass = {cs.CV},
  url          = {https://arxiv.org/abs/2606.08719}
}

@misc{cai2026xkd,
  title        = {{$\mathcal{X}$-KD}: General Experiential Knowledge Distillation for Large Language Models},
  author       = {Yuang Cai and Yuyu Yuan},
  year         = {2026},
  eprint       = {2602.12674},
  archivePrefix = {arXiv},
  primaryClass = {cs.CL},
  url          = {https://arxiv.org/abs/2602.12674}
}

@misc{cai2026effopd,
  title        = {Learning to Foresee: Unveiling the Unlocking Efficiency of On-Policy Distillation},
  author       = {Yuchen Cai and Ding Cao and Liang Lin and Chunxi Luo and Xin Xu and Kai Yang and Weijie Liu and Saiyong Yang and Tianxiang Zhao and Guangzhong Sun and Guiquan Liu and Junfeng Fang},
  year         = {2026},
  eprint       = {2605.11739},
  archivePrefix = {arXiv},
  primaryClass = {cs.CL},
  url          = {https://arxiv.org/abs/2605.11739}
}

@misc{cao2026xopd,
  title        = {{X-OPD}: Cross-Modal On-Policy Distillation for Capability Alignment in Speech {LLM}s},
  author       = {Di Cao and Dongjie Fu and Hai Yu and Siqi Zheng and Xu Tan and Tao Jin},
  year         = {2026},
  eprint       = {2603.24596},
  archivePrefix = {arXiv},
  primaryClass = {eess.AS},
  url          = {https://arxiv.org/abs/2603.24596}
}

@misc{chen2026camopd,
  title        = {Counteraction-Aware Multi-Teacher On-Policy Distillation for General Capability Recovery with Domain Preservation},
  author       = {Tianlei Chen and Jiao Ou and Ziyuan Liu and Ruiming Tang and Jian Liang and Han Li},
  year         = {2026},
  eprint       = {2605.27115},
  archivePrefix = {arXiv},
  primaryClass = {cs.AI},
  url          = {https://arxiv.org/abs/2605.27115}
}

@misc{chen2026fopd,
  title        = {{$\boldsymbol{f}$-OPD}: Stabilizing Long-Horizon On-Policy Distillation with Freshness-Aware Control},
  author       = {Xianwei Chen and Shimin Zhang and Jibin Wu},
  year         = {2026},
  eprint       = {2605.17862},
  archivePrefix = {arXiv},
  primaryClass = {cs.LG},
  url          = {https://arxiv.org/abs/2605.17862}
}

@misc{li2026srpo,
  title        = {Unifying Group-Relative and Self-Distillation Policy Optimization via Sample Routing},
  author       = {Gengsheng Li and Tianyu Yang and Junfeng Fang and Mingyang Song and Mao Zheng and Haiyun Guo and Dan Zhang and Jinqiao Wang and Tat-Seng Chua},
  year         = {2026},
  eprint       = {2604.02288},
  archivePrefix = {arXiv},
  primaryClass = {cs.LG},
  url          = {https://arxiv.org/abs/2604.02288}
}

@misc{li2026safesteer,
  title        = {{SafeSteer}: Localized On-Policy Distillation for Efficient Safety Alignment},
  author       = {Hao Li and Jingkun An and Zijun Song and Pengyu Zhu and Rui Li and Hao Wang and Wendi Feng and Yesheng Liu and Lijun Li and Jin-Ge Yao and Lei Sha},
  year         = {2026},
  eprint       = {2606.02530},
  archivePrefix = {arXiv},
  primaryClass = {cs.AI},
  url          = {https://arxiv.org/abs/2606.02530}
}

@misc{li2026videoopd,
  title        = {{Video-OPD}: Efficient Post-Training of Multimodal Large Language Models for Temporal Video Grounding via On-Policy Distillation},
  author       = {Jiaze Li and Hao Yin and Haoran Xu and Boshen Xu and Wenhui Tan and Zewen He and Jianzhong Ju and Zhenbo Luo and Jian Luan},
  year         = {2026},
  eprint       = {2602.02994},
  archivePrefix = {arXiv},
  primaryClass = {cs.CV},
  url          = {https://arxiv.org/abs/2602.02994}
}

@misc{li2026gear,
  title        = {{GEAR}: Granularity-Adaptive Advantage Reweighting for {LLM} Agents via Self-Distillation},
  author       = {Sijia Li and Yuchen Huang and Zifan Liu and Yanping Li and Jingjing Fu and Li Zhao and Jiang Bian and Ling Zhang and Jun Zhang and Rui Wang},
  year         = {2026},
  eprint       = {2605.11853},
  archivePrefix = {arXiv},
  primaryClass = {cs.LG},
  url          = {https://arxiv.org/abs/2605.11853}
}

@misc{li2026cliff,
  title        = {The Extrapolation Cliff in On-Policy Distillation of Near-Deterministic Structured Outputs},
  author       = {Xin Li and Hao Jiang and Annan Wang and Yichi Zhang and Chau Yuen},
  year         = {2026},
  eprint       = {2605.08737},
  archivePrefix = {arXiv},
  primaryClass = {cs.LG},
  url          = {https://arxiv.org/abs/2605.08737}
}

@misc{li2026vpd,
  title        = {Learning from Language Feedback via Variational Policy Distillation},
  author       = {Yang Li and Erik Nijkamp and Semih Yavuz and Shafiq Joty},
  year         = {2026},
  eprint       = {2605.15113},
  archivePrefix = {arXiv},
  primaryClass = {cs.LG},
  url          = {https://arxiv.org/abs/2605.15113}
}

@misc{li2026rethinkopd,
  title        = {Rethinking On-Policy Distillation of Large Language Models: Phenomenology, Mechanism, and Recipe},
  author       = {Yaxuan Li and Yuxin Zuo and Bingxiang He and Jinqian Zhang and Chaojun Xiao and Cheng Qian and Tianyu Yu and Huan-ang Gao and Wenkai Yang and Zhiyuan Liu and Ning Ding},
  year         = {2026},
  eprint       = {2604.13016},
  archivePrefix = {arXiv},
  primaryClass = {cs.LG},
  url          = {https://arxiv.org/abs/2604.13016}
}

@misc{li2026fireopd,
  title        = {Filter, Then Reweight: Rethinking Optimization Granularity in On-Policy Distillation},
  author       = {Yuying Li and Leqi Zheng and Yongzi Yu and Wenrui Zhou and Xuchang Zhong and Xing Hu and Jing Jin and Hangjie Yuan and Tao Feng},
  year         = {2026},
  eprint       = {2606.02684},
  archivePrefix = {arXiv},
  primaryClass = {cs.LG},
  url          = {https://arxiv.org/abs/2606.02684}
}

@misc{liang2026orbit,
  title        = {{ORBIT}: On-policy Exploration-Exploitation for Controllable Multi-Budget Reasoning},
  author       = {Kun Liang and Clive Bai and Xin Xu and Chenming Tang and Sanwoo Lee and Weijie Liu and Saiyong Yang and Yunfang Wu},
  year         = {2026},
  eprint       = {2601.08310},
  archivePrefix = {arXiv},
  primaryClass = {cs.LG},
  url          = {https://arxiv.org/abs/2601.08310}
}

@misc{wang2026teachability,
  title        = {Not All Disagreement Is Learnable: Token Teachability in On-Policy Distillation},
  author       = {Yuanyi Wang and Su Lu and Yanggan Gu and Pengkai Wang and Yifan Yang and Zhaoyi Yan and Congkai Xie and Jianmin Wu and Hongxia Yang},
  year         = {2026},
  eprint       = {2605.26844},
  archivePrefix = {arXiv},
  primaryClass = {cs.LG},
  url          = {https://arxiv.org/abs/2605.26844}
}

@misc{wu2026lightning,
  title        = {Lightning {OPD}: Efficient Post-Training for Large Reasoning Models with Offline On-Policy Distillation},
  author       = {Yecheng Wu and Song Han and Hai Cai},
  year         = {2026},
  eprint       = {2604.13010},
  archivePrefix = {arXiv},
  primaryClass = {cs.LG},
  url          = {https://arxiv.org/abs/2604.13010}
}

@misc{xin2026kltrap,
  title        = {Escaping the {KL} Agreement Trap in On-Policy Distillation},
  author       = {Haoran Xin and Anhao Zhao and Ying Sun and Jin Li and Xiaoyu Shen and Hui Xiong},
  year         = {2026},
  eprint       = {2606.09471},
  archivePrefix = {arXiv},
  primaryClass = {cs.LG},
  url          = {https://arxiv.org/abs/2606.09471}
}

@misc{xing2026tropd,
  title        = {Trust Region On-Policy Distillation},
  author       = {Xingrun Xing and Haoqing Wang and Boyan Gao and Ziheng Li and Yehui Tang},
  year         = {2026},
  eprint       = {2606.01249},
  archivePrefix = {arXiv},
  primaryClass = {cs.LG},
  url          = {https://arxiv.org/abs/2606.01249}
}

@misc{xiong2026ovd,
  title        = {{OVD}: On-policy Verbal Distillation},
  author       = {Jing Xiong and Hui Shen and Shansan Gong and Yuxin Cheng and Jianghan Shen and Chaofan Tao and Haochen Tan and Haoli Bai and Lifeng Shang and Ngai Wong},
  year         = {2026},
  eprint       = {2601.21968},
  archivePrefix = {arXiv},
  primaryClass = {cs.CL},
  url          = {https://arxiv.org/abs/2601.21968}
}

@misc{xu2026sgopd,
  title        = {{SG-OPD}: Sign-Gated On-Policy Distillation via Sign-Consistency Gating and Phased Teacher Sampling},
  author       = {Haoran Xu and Hongyu Wang and Yifei Gao and Jiaze Li and Xiaofeng Zhang and Xiaosong Yuan},
  year         = {2026},
  eprint       = {2606.09304},
  archivePrefix = {arXiv},
  primaryClass = {cs.CL},
  url          = {https://arxiv.org/abs/2606.09304}
}
\bibliographystyle{iclr2027_conference}

\newpage
\appendix

\section{More Related Work}
\label{sec:related_work}

\paragraph{On-policy language model distillation.}
Knowledge distillation for generative language models~\citep{zhong2024revisiting,kim2024promptkd,zhang2024dualspace,lee2024mentorkd,riviere2024gemma2,boizard2024uld,busbridge2025scaling,wu2025akl,zhang2025aligndistil} has increasingly moved
from matching teacher outputs on fixed data toward training on the student's
own generations~\citep{ko2024distillm,zhou2024distillspec,ye2025gad,fan2025ketchup,xu2025kdrl,xu2025rlkd,bousselham2025vold}. MiniLLM~\citep{gu2024minillm} studies reverse KL as a
distillation objective for autoregressive language models, while
GKD~\citep{agarwal2024onpolicy} develops on-policy distillation by applying
teacher feedback to student-generated trajectories, reducing the mismatch
between training and inference distributions. Recent work has further extended
on-policy distillation to settings such as context distillation and diffusion
language models~\citep{ye2026opcd,ren2026topd,song2026survey,sun2026easyopd,agrawal2026distil,byeon2026tutor,cai2026imagineopd,cai2026xkd,cai2026effopd,cao2026xopd,li2026videoopd,liang2026orbit,xiong2026ovd}. These methods primarily study
how teacher supervision should be obtained or applied on on-policy trajectories~\citep{armandpour2026unmasking,chen2026camopd,chen2026fopd,li2026srpo,li2026gear,li2026cliff,li2026vpd,li2026rethinkopd,wu2026lightning,xin2026kltrap,li2026fireopd,wang2026teachability,xing2026tropd,xu2026sgopd}.
Our work instead focuses on the computational realization of token-level
full-vocabulary supervision: we ask whether the complete reverse-KL correction
can be preserved without keeping the complete vocabulary differentiable.

\paragraph{Sparse vocabulary approximation in distillation.}
A complementary line of work reduces the cost of distribution-level
distillation by restricting or sampling the vocabulary used for supervision~\citep{anshumann2025sparse,huang2026tailaware,li2026safesteer}.
Sparse Logit Sampling~\citep{anshumann2025sparse} shows that naively retaining
only a small set of teacher logits can introduce bias, and uses importance
sampling to recover the dense distillation gradient in expectation.
Recent TopK OPD methods further show that renormalizing teacher and student
distributions on a restricted support can discard useful tail information, and
augment the retained support to better represent that missing mass
\citep{huang2026tailaware}. SparseOPD takes a different route: it does not
approximate the teacher--student comparison before the full correction is
formed. Instead, it first constructs the deterministic full-vocabulary
correction from the complete distributions, and only then compresses the
correction for sparse differentiation. This separation lets support selection
depend directly on correction magnitude and preserves aggregate positive and
negative correction mass, while limiting backpropagation to a small set of
output-head entries.

\section{Additional Experimental Details}
\label{app:experimental_details}
\suppressfloats[t]

This section specifies the data splits, student adaptation, optimization,
and evaluation protocols used for the main accuracy experiments, followed
by a separate efficiency study.

\subsection{Training Data and Model Adaptation}
\label{app:data_adaptation}

\paragraph{Training splits.}
Mathematics uses the complete training split of
\href{https://huggingface.co/datasets/guanning-ai/dapo14k}{DAPO14k}.
For chemistry, we use the released seed-42 split of the Chemistry L3
multiple-choice questions in SciKnowEval V2: a fixed 90\%/10\% partition
into training and held-out questions.
Multimodal training uses the original
\href{https://huggingface.co/datasets/MMR1/MMR1-Math-RL-Data-v0}{MMR1-Math-RL-Data-v0}
release. Its validation split is excluded from training.
Table~\ref{tab:task_training} gives the data usage and task-specific settings.

\paragraph{Student adaptation.}
Each method starts from the same student checkpoint within a model--task pair.
We train low-rank adaptation (LoRA) parameters~\citep{hu2021lora}.
For multimodal students, the vision tower remains frozen, and no bias
parameters are trained.
Training hardware, per-run training time, and shared hyperparameters are
listed in Table~\ref{tab:shared_training}.

\paragraph{Prompting.}
All tasks use the student model's native Qwen chat template with thinking
mode disabled. Mathematics prompts request a step-by-step solution and
a boxed final answer. Chemistry uses a chemistry-expert instruction, with
the same prompt construction at training and evaluation.

\begin{table}[!t]
  \centering
  \setlength{\abovecaptionskip}{5pt}
  \setlength{\belowcaptionskip}{4pt}
  \caption{Shared settings for the main accuracy experiments.}
  \label{tab:shared_training}
  \small
  \renewcommand{\arraystretch}{1.12}
  \begin{tabularx}{\linewidth}{@{}l>{\raggedright\arraybackslash}X@{}}
    \toprule
    Setting & Value \\
    \midrule
    Optimizer & AdamW~\citep{loshchilov2019adamw} \\
    Initial learning rate / weight decay & $10^{-5}$ / $0$ \\
    Maximum gradient norm & $1.0$ \\
    Training hardware & $8$ NVIDIA H20 GPUs \\
    Global batch size & $64$ trajectories \\
    Training time per run & $1.8$--$4$ hours \\
    Training completions per prompt & $1$ \\
    Precision / gradient checkpointing & BF16 / enabled \\
    LoRA rank / scaling parameter & $64$ / $128$ \\
    Maximum prompt length & $1{,}024$ tokens \\
    Rollout temperature / top-$p$ & $1.0$ / $1.0$ \\
    Rollout top-$k$ restriction & Disabled \\
    Training seed & $42$ \\
    Rollout engine / tensor parallelism & Colocated vLLM / $1$ per rank \\
    Attention implementation & FlashAttention-2 \\
    \bottomrule
  \end{tabularx}
\end{table}

\begin{table}[!t]
  \centering
  \setlength{\abovecaptionskip}{5pt}
  \setlength{\belowcaptionskip}{4pt}
  \caption{Task-specific training settings.}
  \label{tab:task_training}
  \small
  \setlength{\tabcolsep}{4pt}
  \renewcommand{\arraystretch}{1.14}
  \begin{tabularx}{\linewidth}{@{}>{\raggedright\arraybackslash}p{0.29\linewidth}
      *{3}{>{\raggedright\arraybackslash}X}@{}}
    \toprule
    Setting & Mathematics & Chemistry & Multimodal \\
    \midrule
    Student sizes & 1.7B, 4B & 1.7B, 4B & 2B, 4B \\
    Training questions & Full DAPO14k train split & $1{,}890$ & $5{,}782$ \\
    Training epochs & $1$ & $3$ & $1$ \\
    Maximum response tokens & $2{,}048$ & $512$ & $2{,}048$ \\
    Distillation temperature & $1.0$ & $1.0$ & $1.0$ \\
    \bottomrule
  \end{tabularx}
\end{table}

\subsection{Optimization and Distillation}
\label{app:optimization}

\paragraph{Training implementation.}
Text experiments use Accelerate with DeepSpeed ZeRO-2 and CPU optimizer
offloading. We use the MiniLLM framework to implement on-policy
distillation (OPD) training.

\paragraph{Distillation protocol.}
For every method, the student generates one on-policy completion and the
frozen teacher scores that same completion by teacher forcing.
Full Vocabulary computes the full-vocabulary reverse Kullback--Leibler
divergence from student to teacher.
TopK uses the student's Top-20 support and renormalizes both distributions
within that support. Sampled Token uses the difference between the sampled
action's student and teacher log-probabilities as a score-function
coefficient, without a baseline.
SparseOPD uses a batch-global budget of $20N$, where $N$ is the number of
valid response positions, with at most 32 retained coordinates per sign
at each position. Its sparse output-head row block size is 32, and its
full-vocabulary planning chunk size is 4,096.

\subsection{Evaluation Protocol}
\label{app:evaluation_protocol}

Table~\ref{tab:eval_decoding} gives the decoding settings shared by all
methods within each task.
For mathematics and chemistry, Mean@8 averages the binary correctness
of eight independently generated answers for each question, then averages
over questions. 
We keep answer extraction and grading consistent across methods within
each benchmark.
For multimodal benchmarks, we follow the LMMs-Eval evaluation
protocol~\citep{zhang2024lmmseval}, extracting answers and grading them
by exact matching, with one generation per question. Each evaluation split
is evaluated in full.

\begin{table}[!t]
  \centering
  \setlength{\abovecaptionskip}{5pt}
  \setlength{\belowcaptionskip}{4pt}
  \caption{Evaluation decoding parameters. All tasks disable the model's
  thinking mode and impose no top-$k$ sampling restriction.}
  \label{tab:eval_decoding}
  \small
  \setlength{\tabcolsep}{5pt}
  \renewcommand{\arraystretch}{1.12}
  \begin{tabularx}{\linewidth}{@{}>{\raggedright\arraybackslash}Xccc@{}}
    \toprule
    Setting & Mathematics & Chemistry & Multimodal \\
    \midrule
    Generations per question & $8$ & $8$ & $1$ \\
    Temperature & $0.6$ & $0.6$ & $1.0$ \\
    Top-$p$ & $0.9$ & $0.9$ & $0.9$ \\
    Maximum prompt tokens & $1{,}024$ & $1{,}024$ & $1{,}024$ \\
    Maximum output tokens & $4{,}096$ & $512$ & $2{,}048$ \\
    \bottomrule
  \end{tabularx}
\end{table}

\subsection{Efficiency Measurements}
\label{app:efficiency_measurements}

\paragraph{Experimental design.}
We profile Qwen3-4B-Base with Qwen3-4B as the frozen teacher.
This study uses full-parameter updates, separately from the LoRA accuracy
experiments. Each method processes the same data. Table~\ref{tab:efficiency_protocol} lists the protocol.

\begin{table}[htbp]
  \centering
  \small
  \setlength{\abovecaptionskip}{5pt}
  \setlength{\belowcaptionskip}{4pt}
  \caption{Protocol for the efficiency measurements.}
  \label{tab:efficiency_protocol}
  \renewcommand{\arraystretch}{1.08}
  \begin{tabularx}{\linewidth}{@{}l>{\raggedright\arraybackslash}X@{}}
    \toprule
    Setting & Value \\
    \midrule
    Hardware & Two H200 GPUs \\
    Teacher / student & Qwen3-4B / Qwen3-4B-Base \\
    Updated parameters & All student parameters \\
    Measured completion lengths & 2,048 / 4,096 / 8,192 / 16,384 tokens \\
    Global batch / microbatch per GPU & 64 / 1 sequence \\
    Gradient accumulation & 32 microbatches per GPU \\
    Measured updates per configuration & 1 \\
    TopK / average SparseOPD support budget & 20 / 20 coordinates per valid position \\
    \bottomrule
  \end{tabularx}
\end{table}

\paragraph{Timing boundaries.}
Forward time covers the loss-producing forward calls accumulated over
the update; backward time covers the corresponding autograd calls,
including recomputation and communication but excluding the optimizer.
Each stage is summed within each GPU, and we report the larger of the
two GPU totals. Full Vocabulary evaluates both student and teacher in
the forward stage. SparseOPD evaluates the teacher, computes full-vocabulary
statistics, and constructs its sparse plan beforehand.

\paragraph{Memory definition.}
Let $M_r^{\mathrm{start}}$ be GPU $r$'s Torch allocated memory at step start,
and $M_r(t)$ its allocation during a stage $\phi$. The main table reports
\[
  M_{\mathrm{extra}}^{\phi}
  =\max_r\left\{\max_{t\in\phi}M_r(t)-M_r^{\mathrm{start}}\right\}.
\]
We use the same pre-step baseline for forward and backward, retaining the
memory of tensors saved by the forward pass for backward.
One GiB equals $2^{30}$ bytes. This metric measures a net increase above
existing allocations.

\begin{table}[!t]
  \centering
  \small
  \setlength{\abovecaptionskip}{5pt}
  \setlength{\belowcaptionskip}{4pt}
  \caption{\textbf{Efficiency results for all four methods.}
  Memory columns are total peak Torch allocations during each stage,
  before baseline subtraction; the main table instead reports extra memory.
  Step time covers the complete update window, including vocabulary sparsification planning.
  Bold marks the lowest measured value at each length, including ties.}
  \label{tab:efficiency_all_methods}
  \renewcommand{\arraystretch}{1.07}
  \fuseresultstable{llrrrrr}{14}{
    \toprule
    \multirow{2}{*}{Length} & \multirow{2}{*}{Method}
    & \multicolumn{2}{c}{Forward} & \multicolumn{2}{c}{Backward}
    & \multirow{2}{*}{Step (s)} \\
    \cmidrule(lr){3-4}\cmidrule(lr){5-6}
    & & Time (s) & Peak (GiB) & Time (s) & Peak (GiB) & \\
    \midrule
    2K & Full Vocabulary & 6.25 & 84.82 & 20.49 & 83.46 & 33.31 \\
       & Sampled Token & 6.51 & 82.48 & 20.18 & \textbf{81.55} & 32.86 \\
       & TopK & 5.82 & 81.27 & \textbf{18.64} & \textbf{81.55} & \textbf{30.14} \\
    \rowcolor{fuseours}
       & SparseOPD(Ours) & \textbf{3.30} & \textbf{79.42} & 19.07 & 81.57 & 37.45 \\
    \midrule
    4K & Full Vocabulary & 10.64 & 91.23 & 28.07 & 88.50 & 45.20 \\
       & Sampled Token & 9.83 & 86.42 & 26.11 & 83.97 & 42.23 \\
       & TopK & 9.39 & 83.99 & \textbf{24.25} & \textbf{81.95} & \textbf{40.27} \\
    \rowcolor{fuseours}
       & SparseOPD(Ours) & \textbf{4.96} & \textbf{80.35} & 24.56 & 82.00 & 50.97 \\
    \midrule
    8K & Full Vocabulary & 19.91 & 103.91 & 45.61 & 98.46 & 72.04 \\
       & Sampled Token & 18.79 & 94.28 & 42.60 & 89.40 & 68.05 \\
       & TopK & 18.48 & 89.43 & \textbf{37.58} & 84.76 & \textbf{61.95} \\
    \rowcolor{fuseours}
       & SparseOPD(Ours) & \textbf{9.14} & \textbf{82.29} & 39.18 & \textbf{84.32} & 82.50 \\
    \midrule
        16K & Full Vocabulary & 39.1 & 129.27 & 83.9 & 118.37 & 129.5 \\
    & Sampled Token & 39.8 & 110.02 & 81.8 & 100.25 & 128.9 \\
    & TopK & 39.3 & 100.31 & \textbf{67.1} & 90.37 & 112.5 \\
\rowcolor{fuseours}
    & SparseOPD(Ours) & \textbf{18.6} & \textbf{86.18} & 73.3 & \textbf{88.96} & 142.7 \\

    \bottomrule
  }
\end{table}

\paragraph{Planning cost and implementation.}
At 8K sequence length, SparseOPD reduces peak extra backward memory
by 70.5\% relative to Full Vocabulary with only a 14.5\% increase in
complete-update time.
The time outside forward and backward also includes input preparation
and optimizer work, so it cannot all be attributed to token selection.
Within backward, selected hidden and weight rows must still be materialized,
dense parameter gradients accumulated, and the full Transformer backbone
backpropagated through, which limits the corresponding time reduction.

\subsection{Parameter Sensitivity Analysis}
\label{app:parameter_sensitivity}

\paragraph{Multiple seeds.}
Table~\ref{tab:parameter_sensitivity} evaluates seed and budget sensitivity
on 1.7B mathematical reasoning under a fixed evaluation protocol.
At the default budget, SparseOPD consistently achieves strong average
accuracy across training seeds, with a small standard deviation.
This consistency suggests that the main result is not driven by a
particularly favorable seed.

\paragraph{Support budget.}
We vary the average support budget $K\in\{16,20,24\}$ while fixing the
training seed. SparseOPD maintains strong performance across the tested
settings, with larger budgets improving average accuracy and $K=24$
achieving the best result. MATH500 and Minerva remain relatively stable,
whereas AMC23 benefits more from increased support, indicating that budget
sensitivity varies across benchmarks while overall performance remains robust.

\begin{table}[!t]
    \centering
    \setlength{\abovecaptionskip}{8pt}
    \setlength{\belowcaptionskip}{5pt}
    \caption{\textbf{Parameter sensitivity on 1.7B mathematical reasoning.}
    Scores are Mean@8 accuracy (\%) under the same evaluation protocol;
    Avg. is the unweighted mean over benchmarks. SparseOPD(Ours) reproduces the main
    result ($K=20$, seed 42). The shaded summary reports mean $\pm$ sample
    standard deviation across seeds 42, 43, and 44, computed from the
    reported scores. Budget variants use seed 42.}
    \label{tab:parameter_sensitivity}
    \renewcommand{\arraystretch}{1.13}
    \providecolor{fusesummary}{HTML}{E8F2EC}
    \fuseresultstable{l r r r r >{\columncolor{fuseavg}}r}{12}{
        \toprule
        \textbf{Setting}
        & \multicolumn{1}{c}{MATH500}
        & \multicolumn{1}{c}{Minerva}
        & \multicolumn{1}{c}{AMC23}
        & \multicolumn{1}{c}{AIME25}
        & \multicolumn{1}{c}{\textbf{Avg.}} \\
        \midrule
        \rowcolor{fuseours}
        \textbf{SparseOPD(Ours)}
        & 64.5 & 26.1 & 35.9 & 5.8 & 33.1 \\
        Seed 43 ($K=20$)
        & 65.10 & 25.28 & 34.69 & 5.42 & 32.62 \\
        Seed 44 ($K=20$)
        & 65.70 & 25.60 & 37.19 & 3.75 & 33.06 \\
        \rowcolor{fusesummary}
        Mean $\pm$ std
        & $65.10\!\pm\!0.60$ & $25.66\!\pm\!0.41$
        & $35.93\!\pm\!1.25$ & $4.99\!\pm\!1.09$ & $32.93\!\pm\!0.27$ \\
        \midrule
        $K=16$
        & 64.25 & 26.15 & 31.25 & 5.42 & 31.77 \\
        $K=24$
        & 65.60 & 26.42 & 36.25 & 5.83 & 33.53 \\
        \bottomrule
    }
\end{table}


\definecolor{casequestion}{HTML}{3678A4}
\definecolor{caseanswer}{HTML}{7853A3}

\newtcolorbox{questionbox}[1]{
  enhanced,
  unbreakable,
  colback=casequestion!4!white,
  colframe=casequestion!48!white,
  colbacktitle=casequestion!12!white,
  coltitle=black,
  title={Question #1},
  fonttitle=\bfseries\small,
  fontupper=\small,
  boxrule=0.45pt,
  arc=2pt,
  left=7pt,
  right=7pt,
  top=5pt,
  bottom=5pt,
  before skip=6pt,
  after skip=5pt
}

\newtcolorbox{answerbox}{
  enhanced,
  unbreakable,
  colback=caseanswer!4!white,
  colframe=caseanswer!48!white,
  colbacktitle=caseanswer!12!white,
  coltitle=black,
  title={Standard Answer},
  fonttitle=\bfseries\small,
  fontupper=\small,
  boxrule=0.45pt,
  arc=2pt,
  left=7pt,
  right=7pt,
  top=5pt,
  bottom=5pt,
  before skip=4pt,
  after skip=8pt
}

\clearpage

\subsection{Case Study}

\label{app:case_study}

\subsubsection{Mathematics}

\label{app:case_mathematics}


\noindent\begin{minipage}{\linewidth}
\begin{questionbox}{M1}
For which $n$ is $n^4 + 6n^3 + 11n^2 + 3n + 31$ a perfect square?
\end{questionbox}
\begin{answerbox}
$10$.
\end{answerbox}
\end{minipage}\par


\noindent\begin{minipage}{\linewidth}
\begin{questionbox}{M2}
Determine the smallest prime $p$ such that $2018!$ is divisible by $p^3$, but not divisible by $p^4$.
\end{questionbox}
\begin{answerbox}
$509$.
\end{answerbox}
\end{minipage}\par


\noindent\begin{minipage}{\linewidth}
\begin{questionbox}{M3}
Let $ABCD$ be a unit square in the plane. Points $X$ and $Y$ are chosen independently and uniformly at random on the perimeter of $ABCD$. If the expected value of the area of triangle $\triangle AXY$ can be expressed as $\frac{m}{n}$ for relatively prime positive integers $m$ and $n$, compute $m+n$.
\end{questionbox}
\begin{answerbox}
$113$.
\end{answerbox}
\end{minipage}\par


\noindent\begin{minipage}{\linewidth}
\begin{questionbox}{M4}
Let $a, b, c$ be distinct numbers such that the equations $x^2 + ax + 1 = 0$ and $x^2 + bx + c = 0$ have a common real root, and the equations $x^2 + x + a = 0$ and $x^2 + cx + b = 0$ also have a common real root. Compute the sum $a + b + c$.
\end{questionbox}
\begin{answerbox}
$-3$.
\end{answerbox}
\end{minipage}\par


\noindent\begin{minipage}{\linewidth}
\begin{questionbox}{M5}
Four positive integers $a, b, c, d$ satisfy the condition $a < b < c < d$. Determine the smallest possible value of $d$ such that the arithmetic mean of the numbers $a, b, c$ is half of the arithmetic mean of the numbers $a, b, c, d$.
\end{questionbox}
\begin{answerbox}
$10$.
\end{answerbox}
\end{minipage}\par

\clearpage

\subsubsection{Chemistry}

\label{app:case_chemistry}


\noindent\begin{minipage}{\linewidth}
\begin{questionbox}{C1}
Determine the pH of the solution of 500 mL of 0.1 M
$\mathrm{CH_3COOH}$, 400 mL of 0.2 M $\mathrm{HCl}$, and
300 mL of 0.3 M $\mathrm{Ba(OH)_2}$.

\smallskip
A. 8.92 \quad B. 8.68 \quad C. 1.38 \quad D. 12.62.
\end{questionbox}
\begin{answerbox}
D. 12.62.
\end{answerbox}
\end{minipage}\par


\noindent\begin{minipage}{\linewidth}
\begin{questionbox}{C2}
An unknown compound has a very broad Fourier-transform infrared (FTIR)
absorption near $3000\,\mathrm{cm}^{-1}$ and a strong absorption near
$1700\,\mathrm{cm}^{-1}$. Its proton nuclear magnetic resonance
(${}^1$H NMR) spectrum contains no vinyl-hydrogen signals, but includes
a doublet of triplets of quartets and a doublet of triplets of triplets.
Identify the compound.

\smallskip
\begin{tabular}{@{}ll@{}}
A. & $\mathrm{CH_3C(H)(C_2H_5)C(H)(C_2H_5)CH_2COOH}$ \\
B. & $\mathrm{CH_3CH_2C(H)(C_2H_5)C(H)(C_2H_5)COOH}$ \\
C. & $\mathrm{CH_3C(H)(CH_3)C(H)(CH_3)CH_2COOH}$ \\
D. & $\mathrm{CH_3CH_2C(H)(CH_3)C(H)(CH_3)COOH}$
\end{tabular}
\end{questionbox}
\begin{answerbox}
A. $\mathrm{CH_3C(H)(C_2H_5)C(H)(C_2H_5)CH_2COOH}$.
\end{answerbox}
\end{minipage}\par


\noindent\begin{minipage}{\linewidth}
\begin{questionbox}{C3}
acetic acid is treated with bromine, pyridine, and acetic anhydride with heating, forming product 1.

1 is heated with ethanol and a small amount of sulfuric acid, forming product 2.

2 is treated with sodium cyanide, forming product 3.

3 is then treated with excess sodium hydride and 1,5-dibromopentane, forming final product 4.

how many distinct hydrogen signals will be observable in the ${}^1$H NMR spectrum of 4? (some of them maybe very close in chemical shift and thus not practically distinguishable, but the desired answer is the number of chemically distinct hydrogens)

A. 10\par
B. 12\par
C. 5\par
D. 8\par
\end{questionbox}
\begin{answerbox}
D. 8
\end{answerbox}
\end{minipage}\par

\clearpage

\subsubsection*{Chemistry (continued)}


\noindent\begin{minipage}{\linewidth}
\begin{questionbox}{C4}
A chemical reaction for the synthesis of a product containing $\mathrm{H}^{+}$ ion was proceeding at room temperature and pH 1.  Accidentally an unknown substance was fallen into the running reaction making the rate of the reaction slower for the product formation and the container got hot due to an exothermic reaction. The pH value of the solution changed to 4 after this accidental addition. What can be the possible reason for changing the rate of reaction?

A. The increased volume of the solution\par
B. The increased pressure of the solution\par
C. The increased pH of the solution\par
D. The increased temperature of the solution\par
\end{questionbox}
\begin{answerbox}
C. The increased pH of the solution
\end{answerbox}
\end{minipage}\par


\noindent\begin{minipage}{\linewidth}
\begin{questionbox}{C5}
You are analyzing a small peptidic compound that has been chemically synthesized. The ${}^1$H NMR spectrum of the crude compound appears consistent with the expected molecule, with the exception of two peaks that both correspond to the same alpha-proton. These two peaks have similar chemical shifts and roughly equal integrals (together they integrate to 1H - the expected integral); based on the coupling pattern, you can rule out spin-spin coupling as an explanation for the duplicate peaks. LC-MS analysis of the crude compound at elevated temperature shows two clearly defined peaks of equal intensities. Both peaks have the same mass spectrum, which is consistent with the expected molecule. What is the most likely explanation for these observations?

A. The crude compound exists as a mixture of enantiomers\par
B. 'Double coupling' has occurred during an amide-bond forming reaction\par
C. The crude compound exists as a mixture of diastereoisomers\par
D. The compound is contaminated with a precursor\par
\end{questionbox}
\begin{answerbox}
C. The crude compound exists as a mixture of diastereoisomers
\end{answerbox}
\end{minipage}\par

\clearpage

\subsubsection{Multimodal Geometry}

\label{app:case_multimodal}

\noindent\begin{minipage}{\linewidth}
\begin{questionbox}{G1}
\begin{minipage}[c]{0.57\linewidth}
One side of a square is a diameter of a circle. The length of one side of the square is 5 feet. To the nearest hundredth, what is the probability that a point chosen at random is in the shaded region?
\end{minipage}\hfill
\begin{minipage}[c]{0.39\linewidth}\centering
\includegraphics[width=\linewidth,height=36mm,keepaspectratio]{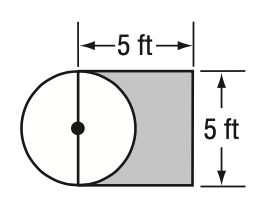}
\end{minipage}
\end{questionbox}
\begin{answerbox}
$0.44$.
\end{answerbox}
\end{minipage}\par
\noindent\begin{minipage}{\linewidth}
\begin{questionbox}{G2}
\begin{minipage}[c]{0.57\linewidth}
In the figure, the vertices of quadrilateral ABCD intersect square EFGH and divide its sides into segments with measures that have a ratio of 1:2. Find the ratio between the areas of ABCD and EFGH.
\end{minipage}\hfill
\begin{minipage}[c]{0.39\linewidth}\centering
\includegraphics[width=\linewidth,height=36mm,keepaspectratio]{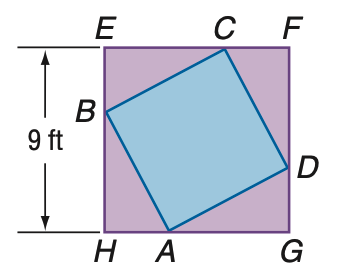}
\end{minipage}
\end{questionbox}
\begin{answerbox}
$5:9$.
\end{answerbox}
\end{minipage}\par
\noindent\begin{minipage}{\linewidth}
\begin{questionbox}{G3}
\begin{minipage}[c]{0.57\linewidth}
A plane travels from Des Moines to Phoenix, on to Atlanta, and back to Des Moines, as shown below. Find the distance in miles from Des Moines to Phoenix if the total trip was 3482 miles.
\end{minipage}\hfill
\begin{minipage}[c]{0.39\linewidth}\centering
\includegraphics[width=\linewidth,height=36mm,keepaspectratio]{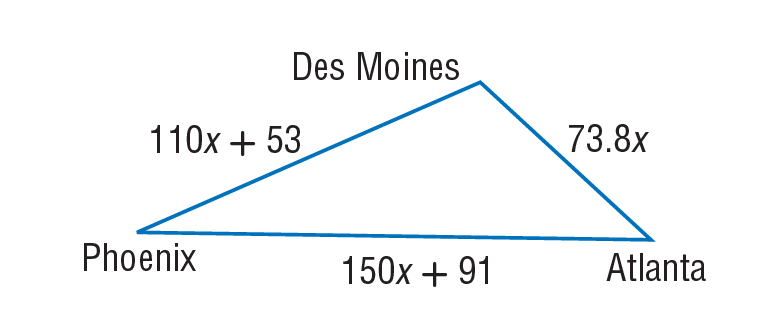}
\end{minipage}
\end{questionbox}
\begin{answerbox}
$1153$.
\end{answerbox}
\end{minipage}\par
\clearpage
\subsubsection*{Multimodal Geometry (continued)}
\noindent\begin{minipage}{\linewidth}
\begin{questionbox}{G4}
\begin{minipage}[c]{0.57\linewidth}
The diagonals of rectangle $A B C D$ each have a length of 56 feet. If $m \angle B A C=42^{\circ}$, what is the length of $\overline{A B}$ to the nearest tenth of a foot?
\end{minipage}\hfill
\begin{minipage}[c]{0.39\linewidth}\centering
\includegraphics[width=\linewidth,height=36mm,keepaspectratio]{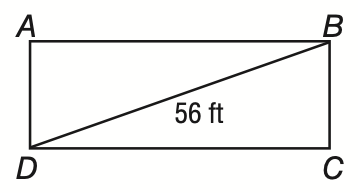}
\end{minipage}
\end{questionbox}
\begin{answerbox}
$41.6$.
\end{answerbox}
\end{minipage}\par
\noindent\begin{minipage}{\linewidth}
\begin{questionbox}{G5}
\begin{minipage}[c]{0.57\linewidth}
$\overrightarrow{BA}$ and $\overrightarrow{BC}$ are opposite rays and $\overrightarrow{BD}$ bisects $\angle ABF$. If $m \angle ABF=3x-8$ and $m \angle ABD=x+14$, find $m \angle ABD$.
\end{minipage}\hfill
\begin{minipage}[c]{0.39\linewidth}\centering
\includegraphics[width=\linewidth,height=36mm,keepaspectratio]{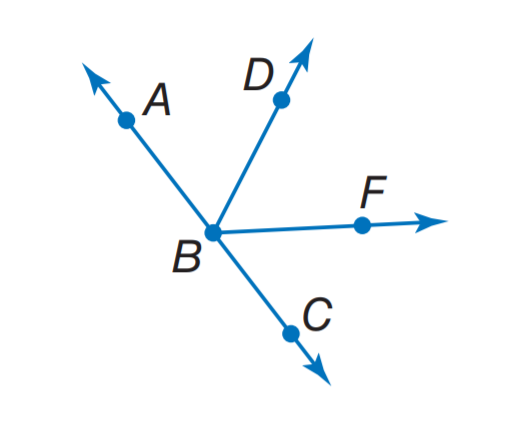}
\end{minipage}
\end{questionbox}
\begin{answerbox}
$50$.
\end{answerbox}
\end{minipage}\par

\end{document}